\documentclass[runningheads]{llncs}

\usepackage{color,xcolor}
\usepackage{epsfig}
\usepackage{graphicx}

\usepackage{adjustbox}
\usepackage{array}
\newcolumntype{?}[1]{!{\vrule width #1}}
\usepackage{booktabs}
\usepackage{colortbl}
\usepackage{float,wrapfig}
\usepackage{hhline}
\usepackage{multirow}
\usepackage{subcaption} 
\usepackage[labelfont={bf},labelsep={period},font={small}]{caption}

\let\llncssubparagraph\subparagraph
\let\subparagraph\paragraph
\usepackage[compact]{titlesec}
\let\subparagraph\llncssubparagraph

\usepackage{amsmath,amsfonts,amssymb}
\usepackage{bm}
\usepackage{nicefrac}
\usepackage{microtype}

\usepackage{changepage}
\usepackage{extramarks}
\usepackage{fancyhdr}
\usepackage{lastpage}
\usepackage{setspace}
\usepackage{soul}
\usepackage{xspace}

\usepackage{url}

\usepackage{algorithm, algorithmic}
\usepackage{enumerate}

\newcolumntype{L}[1]{>{\raggedright\let\newline\\\arraybackslash\hspace{0pt}}m{#1}}
\newcolumntype{C}[1]{>{\centering\let\newline\\\arraybackslash\hspace{0pt}}m{#1}}
\newcolumntype{R}[1]{>{\raggedleft\let\newline\\\arraybackslash\hspace{0pt}}m{#1}}

\newcommand{\fig}[1]{Figure~\ref{#1}}

\newcommand{\tab}[1]{Table~\ref{#1}}

\newcommand{\ignorethis}[1]{}

\def\naive{na\"{\i}ve\xspace}

\makeatletter
\DeclareRobustCommand\onedot{\futurelet\@let@token\@onedot}
\def\@onedot{\ifx\@let@token.\else.\null\fi\xspace}

\def\eg{\emph{e.g}\onedot} 
\def\ie{\emph{i.e}\onedot} 
 
 \def\vs{\emph{vs}\onedot}
 
\def\etal{\emph{et al}\onedot}
\makeatother

\definecolor{citecolor}{RGB}{34,139,34}
\definecolor{mydarkblue}{rgb}{0,0.08,1}
\definecolor{mydarkgreen}{rgb}{0.02,0.6,0.02}
\definecolor{mydarkred}{rgb}{0.8,0.02,0.02}
\definecolor{mydarkorange}{rgb}{0.40,0.2,0.02}
\definecolor{mypurple}{RGB}{111,0,255}
\definecolor{myred}{rgb}{1.0,0.0,0.0}
\definecolor{mygold}{rgb}{0.75,0.6,0.12}
\definecolor{mydarkgray}{rgb}{0.66, 0.66, 0.66}

\def\module{Sparse Point-Voxel Convolution\xspace}
\def\moduleshort{SPVConv\xspace}

\def\cnnshort{SPVCNN\xspace}

\def\modelshort{SPVNAS\xspace}

\def\nas{3D Neural Architecture Search\xspace}
\def\nasshort{3D-NAS\xspace}

\def\ECCVSubNumber{6421}

\title{Searching Efficient 3D Architectures with \\ \module}
\author{Haotian Tang$^{1,*}$ \and Zhijian Liu$^{1,*}$ \and \\ Shengyu Zhao$^{1,2}$ \and Yujun Lin$^{1}$ \and Ji Lin$^{1}$ \and Hanrui Wang$^{1}$ \and Song Han$^{1}$}
\institute{$^{1}$ Massachusetts Institute of Technology \\ $^{2}$ IIIS, Tsinghua University}

\titlerunning{Searching Efficient 3D Architectures with \module} 
\authorrunning{H. Tang$^{*}$, Z. Liu$^{*}$, S. Zhao, Y. Lin, J. Lin, H. Wang, and S. Han} 

\begin{document}

\pagestyle{headings}
\mainmatter

\maketitle

\footnotetext{$*$ indicates equal contributions; order determined by a coin toss.}

\begin{abstract}

Self-driving cars need to understand 3D scenes efficiently and accurately in order to drive safely. Given the limited hardware resources, existing 3D perception models are not able to recognize small instances (\eg, pedestrians, cyclists) very well due to the low-resolution voxelization and aggressive downsampling. To this end, we propose \emph{\module (\moduleshort)}, a lightweight 3D module that equips the vanilla Sparse Convolution with the high-resolution point-based branch. With negligible overhead, this point-based branch is able to preserve the fine details even from large outdoor scenes. To explore the spectrum of efficient 3D models, we first define a flexible architecture design space based on \moduleshort, and we then present \emph{\nas (\nasshort)} to search the optimal network architecture over this diverse design space efficiently and effectively. Experimental results validate that the resulting \modelshort model is fast and accurate: it outperforms the state-of-the-art MinkowskiNet by 3.3\%, ranking \textbf{1\textsuperscript{st}} on the competitive SemanticKITTI leaderboard upon publication. It also achieves \textbf{8$\times$} computation reduction and \textbf{3$\times$} measured speedup over MinkowskiNet still with higher accuracy. Finally, we transfer our method to 3D object detection, and it achieves consistent improvements over the one-stage detection baseline on KITTI.

\end{abstract}
\section{Introduction}

\begin{figure*}[t]
\centering
\begin{subfigure}[t]{0.48\linewidth}
    \centering
    \includegraphics[width=\linewidth]{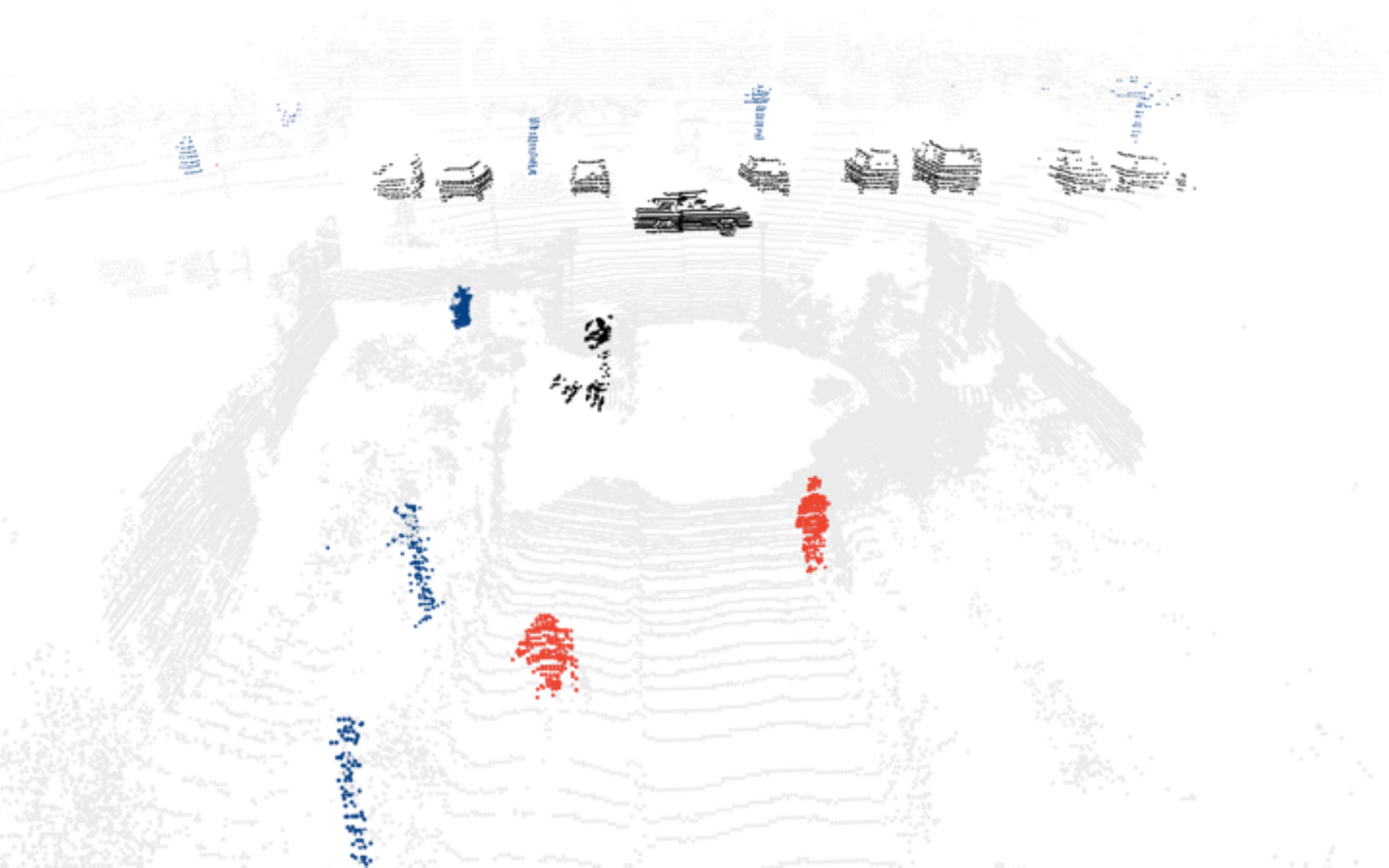}
    \caption{Large 3D Scene}
    \label{fig:teaser:a}
\end{subfigure}
\begin{subfigure}[t]{0.48\linewidth}
    \centering
    \includegraphics[width=\linewidth]{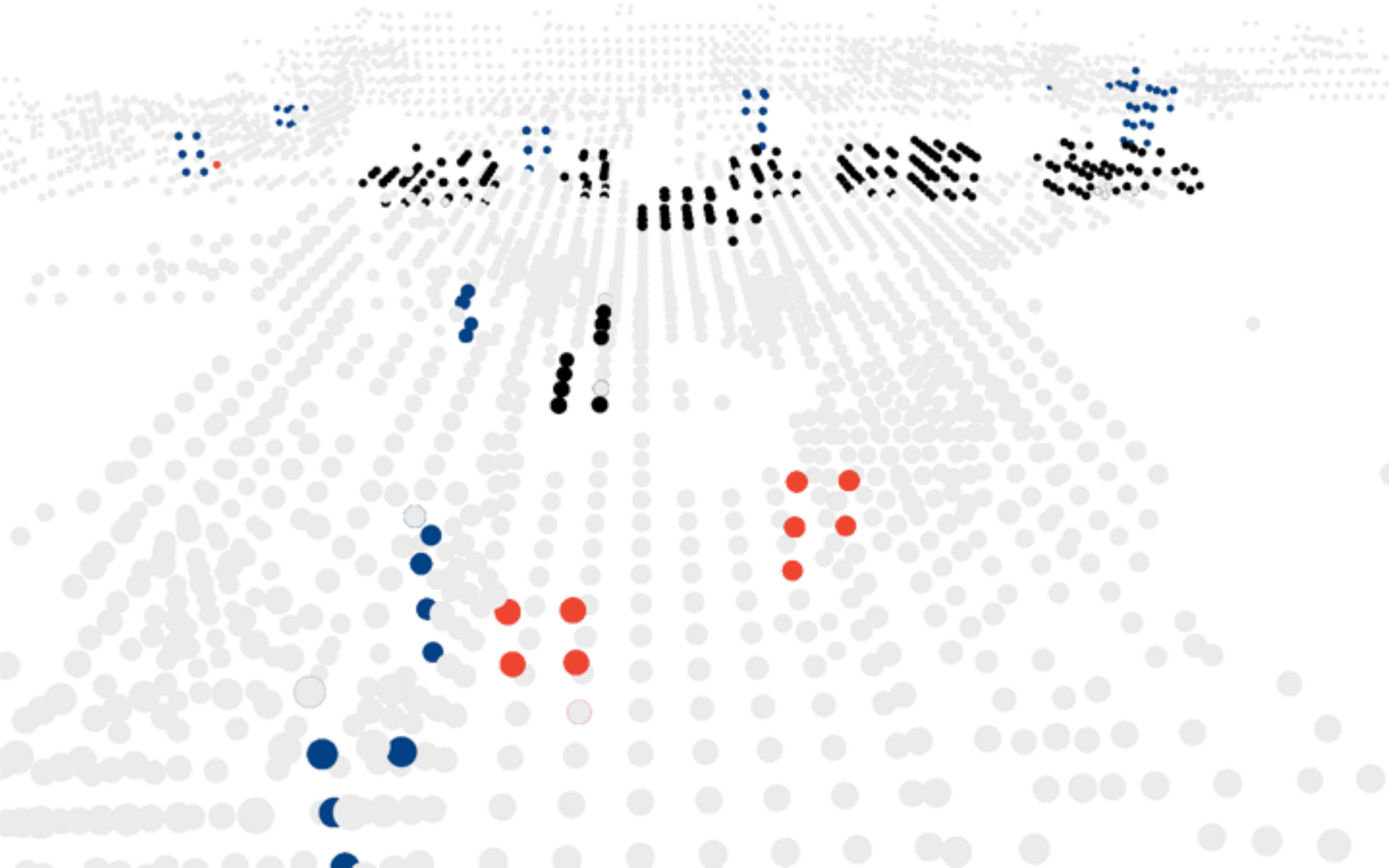}
    \caption{Low Resolution (0.8m)}
    \label{fig:teaser:b}
\end{subfigure}
\caption{Small instances (e.g., pedestrians and cyclists) are hard to be recognized at a low resolution (due to the coarse voxelization or the aggressive downsampling).}
\label{fig:teaser}
\end{figure*}

3D deep learning has received increased attention thanks to its wide applications: \eg, it has been used in LiDAR perception that serves as the eyes of autonomous driving systems to understand the semantics of outdoor scenes. As the safety of the passenger is the top priority of the self-driving cars, 3D perception models are required to achieve high accuracy and low latency at the same time. However, the hardware resources on the self-driving cars are tightly constrained by the form factor (since we do not want a whole trunk of workstations) and heat dissipation. Thus, it is crucial to design efficient and effective 3D neural network models with limited computation resources (\eg, memory).

Researchers have primarily exploited two 3D data representations: point cloud and rasterized voxel grids. As analyzed in Liu~\etal~\cite{liu2019point}, point-based methods~\cite{qi2017pointnet,qi2017pointnet++,li2018pointcnn} waste up to 90\% of their runtime on structuring the irregular data, not on the actual feature extraction. On the other hand, voxel-based methods usually suffer from severe information loss: \ie, the resolution of dense voxels~\cite{maturana2015voxnet,liu2019point} is strictly constrained by the memory; the sparse voxels~\cite{graham20183d,choy20194d} require aggressive downsampling to achieve larger receptive field, leading to low resolution at deeper layers. With low resolution (see \fig{fig:teaser}), multiple points or even multiple small objects might be merged into one grid and become indistinguishable. In this case, small instances (\eg, pedestrians and cyclists) are at a disadvantage compared with large objects (\eg, cars). Therefore, the effectiveness of previous 3D modules is discounted when hardware resources are limited and the resolution is low.

To tackle these problems, we propose a novel 3D module, \emph{\module (\moduleshort)} that introduces a low-cost high-resolution point-based branch to the vanilla Sparse Convolution, which helps to capture the fine details. On top of \moduleshort, we further present \emph{\nas (\nasshort)} to search an efficient 3D model. We incorporate fine-grained channel numbers into the search space to increase the diversity and introduce the progressive depth shrinking to accelerate the training. Experimental results validate that our model is fast and accurate: it outperforms MinkowskiNet by 3.3\% in mIoU with lower latency. It also achieves \textbf{8$\times$} computation reduction and \textbf{3$\times$} measured speedup over MinkowskiNet, while providing higher accuracy. We further transfer our method to KITTI for 3D object detection, and it achieves consistent improvements over the previous one-stage detection baseline.

The contribution of this paper has three aspects:
\begin{enumerate}
    \item We design a lightweight 3D module, \moduleshort, that boosts the performance on small objects, which used to be challenging under limited hardware resource.
    \item We introduce the first AutoML framework for 3D scene understanding, \nasshort, that offers the best 3D model given a specific resource constraint.
    \item Our method outperforms all previous methods with a large margin and ranks 1\textsuperscript{st} on the competitive SemanticKITTI leaderboard\footnote{\url{http://semantic-kitti.org/tasks.html#semseg}} upon publication. It can also be transferred to object detection and achieves consistent improvements.
\end{enumerate}

\section{Related Work}

\subsection{3D Perception Models}

Increased attention has been paid to 3D deep learning, which is important for LiDAR perception in autonomous driving. Early research~\cite{chang2015shapenet,maturana2015voxnet,qi2016volumetric,wang2019voxsegnet,zhou2018voxelnet} relied on the volumetric representation and vanilla 3D convolution to process the 3D data. Due to the \textit{sparse} nature of 3D representation, the \textit{dense} volumetric representation is inherently inefficient, and it also inevitably introduces information loss. Therefore, researchers have proposed to directly learn on the 3D point cloud representation using the symmetric function~\cite{qi2017pointnet}. To improve the neighborhood modeling capability, researchers have defined point-based convolutions on either geometric~\cite{li2018pointcnn,mao2019interpolated,qi2017pointnet++,su2018splatnet,tatarchenko2018tangent,thomas2019kpconv,wu2019pointconv,xu2018spidercnn} or semantic~\cite{wang2018dynamic} neighborhood. There are also 3D models tailored for specific tasks such as detection~\cite{qi2020imvotenet,qi2019deep,qi2018frustum,shi2019pvrcnn,shi2019pointrcnn,shi2019parta2,yan2018second,yang2019std} and instance segmentation~\cite{han2020occuseg,jiang2020pointgroup,lahoud2019multi,yang2019learning} built upon these modules.

Recently, a few researchers started to pay attention to the efficiency aspect of 3D deep learning. Riegler~\etal~\cite{riegler2017octnet}, Wang~\etal~\cite{wang2017cnn,wang2018adaptive} and Lei~\etal~\cite{lei2019octree} proposed to reduce the memory footprint of the volumetric representation using octrees where areas with lower density occupy fewer voxel grids. Liu~\etal~\cite{liu2019point} analyzed the bottlenecks of point-based and voxel-based methods, and proposed Point-Voxel Convolution. Graham~\etal~\cite{graham20183d} and Choy~\etal~\cite{choy20194d} proposed Sparse Convolution to accelerate the volumetric convolution by keeping the activation sparse and skipping the computations in the inactive regions.

\subsection{Neural Architecture Search}

To alleviate the burden of manually designing neural networks~\cite{howard2017mobilenets,sandler2018mobilenetv2,ma2018shufflenet,zhang2018shufflenet,iandola2016squeezenet}, researchers have introduced neural architecture search (NAS) to automatically architect the neural network with high accuracy using reinforcement learning~\cite{zoph2017neural,zoph2018learning} and evolutionary search~\cite{liu2019progressive}. A new wave of research started to design efficient models with neural architecture search~\cite{tan2019mnasnet,wu2019fbnet,tan2019efficientnet} for mobile deployment. However, conventional frameworks require high computation cost and considerable carbon footprint~\cite{strubell2019energy}. To tackle these, researchers have proposed different techniques to reduce the search cost, including differentiable architecture search~\cite{liu2019darts}, path-level binarization~\cite{cai2019proxylessnas}, single-path one-shot sampling~\cite{guo2019single,chen2019detnas,cai2020once}, and weight sharing~\cite{stamoulis2019single,cai2020once,wang2020hat}. Besides, neural architecture search has also been used in compressing and accelerating neural networks, including pruning~\cite{he2018amc,liu2019metapruning,cai2019automl,li2020gan} and quantization~\cite{wang2019haq,guo2019single,wang2020hardware,wang2020apq}. Most of these methods are tailored for 2D visual recognition, which has many well-defined search spaces~\cite{radosavovic2019on}. Lately, researchers have applied neural architecture search to 3D medical image segmentation~\cite{zhu2019vnas,kim2019scalable,yang2019searching,bae2019resource,wong2019segnas3d,yu2020c2fnas} and 3D shape classification~\cite{ma2020auto,li2020sgas}. However, they are not directly applicable to 3D scene understanding since 3D medical data are still in the similar format as 2D images (which are entirely different from 3D scenes), and 3D objects are of much smaller scales than 3D scenes (which makes them less sensitive to the resolution).
\section{\moduleshort: Designing Effective 3D Modules}

We first revisit two recent 3D modules: Point-Voxel Convolution~\cite{liu2019point} and Sparse Convolution~\cite{choy20194d} and analyze their bottlenecks. We observe that both of them suffer from information loss (caused by coarse voxelization or aggressive downsampling) when the memory is constrained. To this end, we introduce \emph{\module (\moduleshort)}, to effectively process the large 3D scene (as in \fig{fig:overview:spvconv}).

\subsection{Point-Voxel Convolution: Coarse Voxelization}

\begin{table}[t]
\setlength{\tabcolsep}{5.5pt}
\small\centering
\begin{tabular}{lcccc}
    \toprule
    & Input & Voxel Size (m) & Latency (ms) & Mean IoU \\
    \midrule
    \multirow{2}{*}{PVConv~\cite{liu2019point}} & Sliding Window & 0.05 & 35640 & -- \\
     & Entire Scene & 0.78 & 146 & 39.0 \\
    \midrule
    \textbf{\moduleshort} (Ours) & Entire Scene & 0.05 & \textbf{85} & \textbf{58.8} \\
    \bottomrule
\end{tabular}
\caption{Point-Voxel Convolution~\cite{liu2019point} is not suitable for large 3D scenes. If processing with sliding windows, the large latency is not affordable for real-time applications. If taking the whole scene, the resolution is too coarse to capture useful information.}
\label{tab:limitation}
\end{table}

Liu~\etal~\cite{liu2019point} proposed Point-Voxel Convolution that represents the 3D input data in points to reduce the memory consumption, while performing the convolutions in voxels to reduce the irregular data access and improve the locality. Specifically, its point-based branch transforms each point individually, and its voxel-based branch convolves over the voxelized input from the point-based branch.

PVCNN (which is built upon Point-Voxel Convolution) can afford at most 128\textsuperscript{3} voxels in its voxel-based branch on a single GPU (with 12 GB of memory). Given a large outdoor scene (with size of 100m$\times$100m$\times$10m), each voxel grid will correspond to a fairly large area (with size of 0.8m$\times$0.8m$\times$0.1m). In this case, the small instances (\eg, pedestrians) will only occupy a few voxel grids (see \fig{fig:teaser}). From such few points, PVCNN can hardly learn any useful information from the voxel-based branch, leading to a relatively low performance (see \tab{tab:limitation}). Alternatively, we can process the large 3D scene piece by piece so that each sliding window is of smaller scale. In order to preserve the fine-grained information (\ie, voxel size is smaller than 0.05m), we have to run PVCNN once for each of the 244 sliding windows. This takes 35 seconds to process a single scene, which is not affordable for most real-time applications (\eg, autonomous driving).

\subsection{Sparse Convolution: Aggressive Downsampling}

Volumetric convolution has always been considered inefficient and prohibitive to be scaled up. Lately, researchers proposed Sparse Convolution~\cite{graham20183d,choy20194d} that skips the non-activated regions to significantly reduce the memory consumption. More specifically, it first finds all active synapses (denoted as \textit{kernel map}) between the input and output points; it then performs the convolution based on this kernel map. To keep the activation sparse, it only considers these output points that also belong to the input. We refer the readers to Choy~\etal~\cite{choy20194d} for more details.

As such, Sparse Convolution can afford a much higher resolution than the vanilla volumetric convolution. However, the network cannot be very deep due to the limited computation resource. As a result, the network has to downsample very aggressively in order to achieve a sufficiently large receptive field, which is very lossy. For instance, the state-of-the-art MinkowskiNet~\cite{choy20194d} gradually applies four downsampling layers to the input point cloud, after which, the voxel size will become $0.05 \times 2^4 = 0.8$m. Similar to Point-Voxel Convolution, this resolution is too coarse to capture the small instances (see \fig{fig:semantickitti:visualizations}).

\subsection{Solution: \module}

\begin{figure*}[t]
\centering
\includegraphics[width=\linewidth]{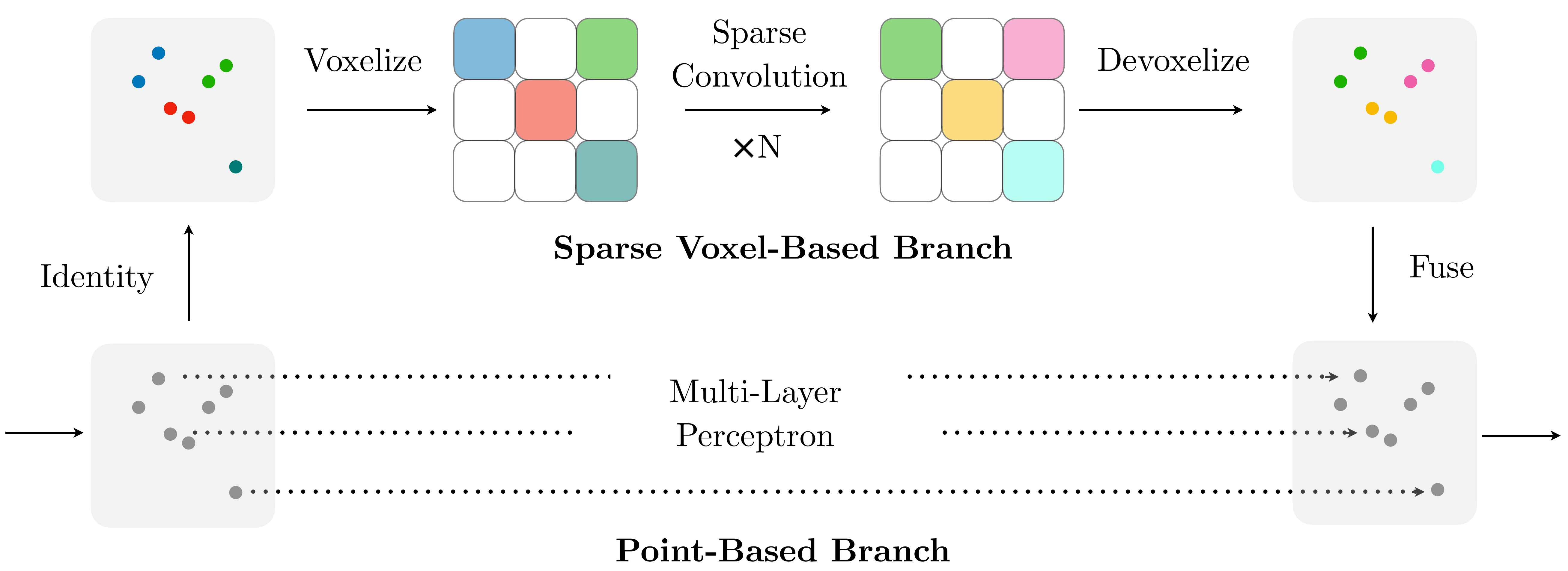}
\caption{Overview of \module (\moduleshort): it equips the sparse voxel-based branch with a lightweight, high-resolution point-based branch which can capture fine details in large scenes.}
\label{fig:overview:spvconv}
\end{figure*}

In order to overcome the limitations of both modules, we present \module (\moduleshort) in \fig{fig:overview:spvconv}: the point-based branch always keeps the high-resolution representation, and the sparse voxel-based branch applies Sparse Convolution to model across different receptive field size. Two branches communicate at a negligible cost through sparse voxelization and devoxelization.

\paragraph{Data Representation.}

Our \module operates on:
\begin{itemize}
     \item sparse voxelized tensor $\bm{S}= (\{(\bm{p}^s_m, \bm{f}^s_m)\}, v)$, where $\bm{p}^s_m = (x^s_m, y^s_m, z^s_m)$ is the 3D coordinate and $\bm{f}^s_m$ is the feature vector of the $m$\textsuperscript{th} nonzero grid, and $v$ is the voxel size (\ie, side length) for one grid in the current layer;
     \item point cloud tensor $\bm{T}= \{(\bm{p}^t_k, \bm{f}^t_k)\}$, where $\bm{p}_k = (x_k, y_k, z_k)$ is the 3D coordinate and $\bm{f}_k$ is feature vector of $k$\textsuperscript{th} point.
\end{itemize}

\paragraph{Sparse Voxelization.}

In the upper sparse voxel-based branch, we first transform the high-resolution point cloud tensor $\bm{T}$ to a sparse tensor $\bm{S}$:
\begin{align}
  \hat{\bm{p}}^t_k = (\hat{x}^t_k, \hat{y}^t_k, \hat{z}^t_k) = (\text{floor}(x^t_k / v), \text{floor}(y^t_k / v), \text{floor}(z^t_k / v)),\\
 \bm{f}^s_{m} = \frac{1}{N_m} \sum_{k=1}^n \mathbb{I}[\hat{x}^t_k = x^s_m,  \hat{y}^t_k = y^s_m, \hat{z}^t_k = z^s_m]\cdot \bm{f}^t_{k},
\label{eqn:sparsevoxelize}
\end{align}
where $\mathbb{I}[\cdot]$ is the binary indicator of whether $\hat{\bm{p}}^t_k$ belongs to the voxel grid $\bm{p}^s_m$, and $N_m$ is the normalization factor (\ie, the number of points that fall into the $m$\textsuperscript{th} nonzero voxel grid). Such formulation, however, requires $\mathcal{O}(mn)$ complexity where $m = |\bm{S}|$ and $n = |\bm{T}|$. With typical values of $m, n$ at the order of $10^5$, the naive implementation is impractical for real-time applications.

To this end, we propose to use the GPU hash table to accelerate the sparse voxelization and devoxelization. Specifically, we first construct a hash table for all activated points in the sparse voxelized tensor $\bm{S}$, which can be completed in $\mathcal{O}(n)$ time. After that, we iterate over all points in $\bm{T}$, and for each point, we use its voxelized coordinate as the key to query the corresponding index in the sparse voxelized tensor. As the lookup over the hash table requires $\mathcal{O}(1)$ time~\cite{pagh2001cuckoo}, this query step will in total take $\mathcal{O}(m)$ time. Therefore, the total time of coordinate indexing will be reduced from $\mathcal{O}(mn)$ to $\mathcal{O}(m+n)$. 

\paragraph{Feature Aggregation.}

We then perform the neighborhood feature aggregation on the sparse voxelized tensor using a sequence of residual Sparse Convolution blocks~\cite{choy20194d}. We parallelize the kernel map operation in Sparse Convolution on GPU with the same hash table implementation as in sparse voxelization, which offers 1.3$\times$ speedup over the implementation from Choy~\etal~\cite{choy20194d}. Note that both our method and the baseline have been upgraded to this accelerated implementation.

\paragraph{Sparse Devoxelization.}

With the aggregated features (which are in the form of sparse tensors), we transform them back to the point-based representation so that the information from both branches can be fused together. Similar to Liu~\etal~\cite{liu2019point}, we choose to interpolate each point's feature with its 8 neighbor voxel grids using trilinear interpolation instead of the \naive nearest interpolation.

\paragraph{Point Transformation and Feature Fusion.}

In the lower point-based branch, we directly apply an MLP on each point to extract individual point features. We then fuse the outputs of two branches with an addition to combine the complementary information provided. Compared with the vanilla Sparse Convolution, MLP layers only cost little computation overhead (4\% in terms of \#MACs) but introduce important fine details into the information flow (see \fig{fig:analysis:attention}).

\section{\nasshort: Searching Efficient 3D Architectures}

\begin{figure*}[t]
\centering
\includegraphics[width=\linewidth]{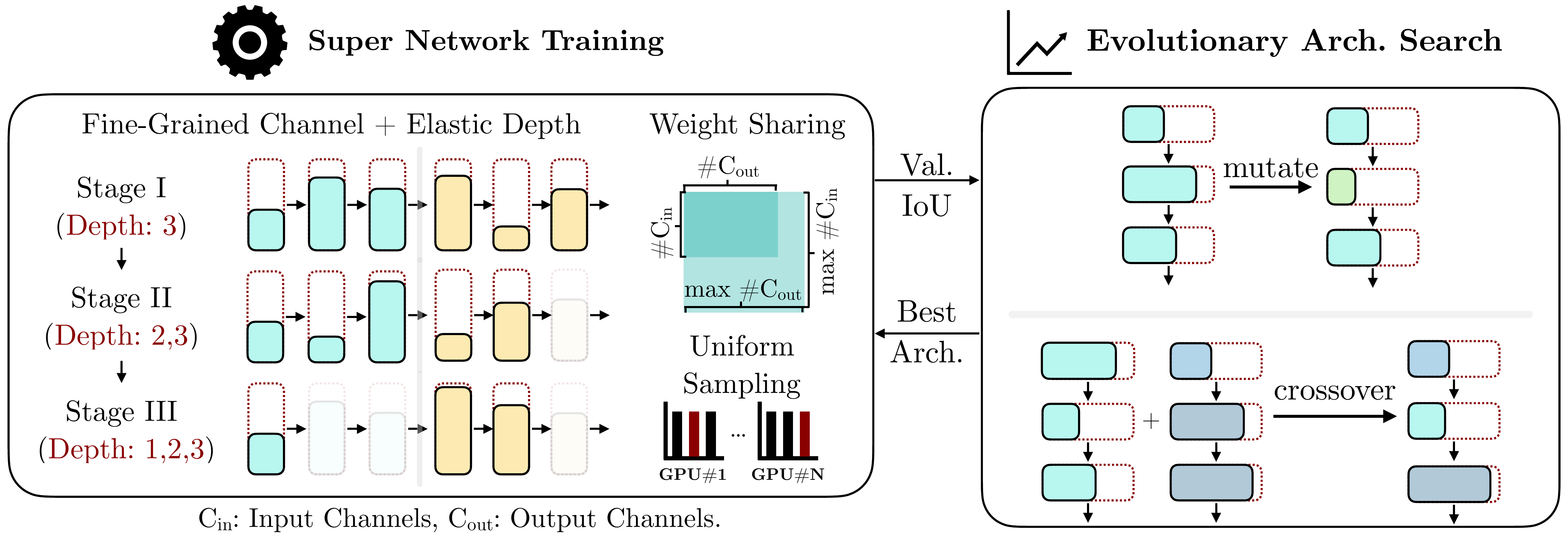}
\caption{Overview of \nas (\nasshort): we first train a super network composed of multiple \moduleshort's, supporting fine-grained channel numbers and elastic network depths. Then, we perform the evolutionary architecture search to obtain the best candidate model under a given computation constraint.}
\label{fig:overview:3dnas}
\end{figure*}

Even with our module, designing an efficient neural network is still challenging. We need to carefully adjust the network architecture (\eg, channel numbers and kernel sizes of all layers) to meet the constraints for real-world applications (\eg, latency, energy, and accuracy). To this end, we introduce \emph{\nas (\nasshort)}, to automatically design efficient 3D models (as in \fig{fig:overview:3dnas}).

\subsection{Design Space}

The performance of neural architecture search is greatly impacted by the design space quality. In our search space, we incorporate fine-grained channel numbers and elastic network depths; however, we do not support different kernel sizes.

\paragraph{Fine-grained Channel Numbers.}

The computation cost increases quadratically with the number of channels; therefore, the channel number selection has a large influence on the network efficiency. Most existing neural architecture frameworks~\cite{cai2019proxylessnas} only support the coarse-grained channel number selection: \eg, searching the expansion ratio of the ResNet/MobileNet blocks over a few (2-3) choices. In this case, only intermediate channel numbers of the blocks can be changed; while the input and output channel numbers will still remain the same. Empirically, we observe that this limits the variety of the search space. To this end, we enlarge the search space by allowing all channel numbers to be selected from a large collection of choices (with size of $O(n)$). This fine-grained channel number selection largely increase the number of candidates for each block: \eg, from constant (2-3) to $\mathcal{O}(n^2)$ for a block with two consecutive convolutions.

\paragraph{Elastic Network Depths.}

We support different network depth in our design space. For 3D CNNs, reducing the channel numbers alone cannot achieve significant measured speedup, which is drastically different from 2D CNNs. For example, by shrinking all channel numbers in MinkowskiNet~\cite{choy20194d} by $4\times$ and $8\times$, the number of MACs will be reduced to 7.5 G and 1.9 G, respectively. However, although the number of MACs is drastically reduced, their measured latency on the GPU is very similar: 105 ms and 96 ms (measured on a single GTX1080Ti GPU). This suggests that scaling down the number of channels alone cannot offer us with very efficient models, even though the number of MACs is very small. This might be because 3D modules are usually more memory-bounded than 2D modules; the number of MACs decreases quadratically with channel number, while memory decreases linearly. Motivated by this, we choose to incorporate the elastic network depth into our design space so that these layers with very small computation (and large memory cost) can be removed and merged into their neighboring layers.

\paragraph{Small Kernel Matters.}

Kernel sizes are usually included into the search space of 2D CNNs. This is because a single convolution with larger kernel size can be more efficient than multiple convolutions with smaller kernel sizes on GPUs. However, it is not the case for the 3D CNNs. From the computation perspective, a single 2D convolution with kernel size of 5 requires only 1.4$\times$ more MACs than two 2D convolutions with kernel sizes of 3; while a single 3D convolution with kernel size of 5 requires 2.3$\times$ more MACs than two 3D convolutions with kernel sizes of 3 (if applied to dense voxel grids). This larger computation cost makes it less suitable to use large kernel sizes in 3D CNNs. Furthermore, the computation overhead of 3D modules is also related to the kernel sizes. For example, Sparse Convolution~\cite{graham20183d,choy20194d} requires $\mathcal{O}(k^3n)$ time to build the kernel map, where $k$ is the kernel size and $n$ is the number of points, which indicates that its cost grows cubically with respect to the kernel size. Based on these reasons, we decide to keep the kernel size of all convolutions to be 3 and do not allow the kernel size to change in our search space. Even with the small kernel size, we can still achieve a large receptive field by changing the network depth, which can achieve the same effect as changing the kernel size.

\subsection{Training Paradigm}

Searching over a fine-grained design space is very challenging as it is impossible to train every sampled candidate network from scratch~\cite{tan2019mnasnet}. Motivated by Guo~\etal~\cite{guo2019single}, we incorporate all candidate networks into a single super network, and after training this super network once, each candidate network can be directly extracted with inherited weights. As such, the total training cost can be reduced from $\mathcal{O}(n)$ to $\mathcal{O}(1)$, where $n$ is the number of candidate networks.

\paragraph{Uniform Sampling.}

At each training iteration, we randomly sample a candidate network from the super network: randomly select the channel number for each layer, and then randomly select the network depth (\ie the number of blocks to be used) for each stage. The total number of candidate networks to be sampled during training is very limited; therefore, we choose to sample different candidate networks on different GPUs and average their gradients at each step so that more candidate networks can be sampled. For 3D, this is more critical because the 3D datasets usually contain fewer training samples than the 2D datasets: \eg 20K on SemanticKITTI~\cite{behley2019semantickitti} \vs 1M on ImageNet~\cite{deng2009imagenet}.

\paragraph{Weight Sharing.}

As the number of candidate networks is enormous, every candidate network will only be optimized for a small fraction of the total schedule. Therefore, uniform sampling alone is not enough to train all candidate networks sufficiently (\ie, achieving the same level of performance as being trained from scratch). To tackle this, we adopt the weight sharing technique so that every candidate network can be optimized at each iteration even if it is not sampled. Specifically, given the input channel number $C_{\text{in}}$ and output channel number $C_{\text{out}}$ of each convolution layer, we simply index the first $C_{\text{in}}$ and $C_{\text{out}}$ channels from the weight tensor accordingly to perform the convolution~\cite{guo2019single}. For each batch normalization layer, we similarly crop the first $c$ channels from the weight tensor based on the sampled channel number $c$. Finally, with the sampled depth $d$ for each stage, we choose to keep the first $d$ layers, instead of randomly sampling $d$ of them. This ensures that each layer will always correspond to the same depth index within the stage.

\paragraph{Progressive Depth Shrinking.}

Suppose we have $n$ stages, each of which has $m$ different depth choices from 1 to $m$. If we sample the depth $d_k$ for each stage $k$ randomly, the expected total depth of the network will be
$\mathbb{E}[d] = \sum_{k=1}^n \mathbb{E}[d_k] = n\times(m+1)/2$,
which is much smaller than the maximum depth $nm$. Furthermore, the probability of the largest candidate network (with the maximum depth) being sampled is extremely small: $m^{-n}$. Therefore, the largest candidate networks are poorly trained due to the small possibility of being sampled. To this end, we introduce progressively depth shrinking to alleviate this issue. We divide the training epochs into $m$ segments for $m$ different depth choices. During the $k^{\text{th}}$ training segment, we only allow the depth of each stage to be selected from $m-k+1$ to $m$. This is essentially designed to enlarge the search space gradually so that these large candidate networks can be sampled more frequently.

\subsection{Search Algorithm}

After the super network is fully trained, we use evolutionary architecture search to find the best architectures under a certain resource constraint.

\paragraph{Resource Constraints.}

We use the number of MACs as the resource constraint. For 3D CNNs, the number of MACs cannot be simply determined by the input size and the network architecture: \eg, Sparse Convolution only performs the computation over the active synapses; therefore, its computation is also determined by the input sparsity pattern. To address this, we first estimate the average kernel map size over the entire dataset for each convolution layer, and we can then measure the number of MACs based on these statistics.

\paragraph{Evolutionary Search.}

We automate the architecture search with the evolutionary algorithm~\cite{guo2019single}. We initialize the starting population with $n$ randomly sampled candidate networks. At every iteration, we evaluate all candidate networks in the population and select the $k$ models with the highest mIoUs (\ie, the fittest individuals). The population for the next iteration is then generated with $(n/2)$ mutations and $(n/2)$ crossovers. For each mutation, we randomly select one among the top-$k$ candidates and alter each of its architectural parameters (\eg, channel numbers, network depths) with a pre-defined probability; for each crossover, we select two from the top-$k$ candidates and generate a new model by fusing them together randomly. Finally, the best model is obtained from the population of the last iteration. During the evolutionary search, we ensure that all the candidate networks in the population always meet the given resource constraint (we will resample another candidate network until the resource constraint is satisfied).
\section{Experiments}

Based on our lightweight 3D module, we first manually construct our backbone network (denoted as \cnnshort). Then, we leverage our neural architecture search framework to explore the best 3D model (denoted as \modelshort). We provide more implementation details in the appendix. Evaluated on 3D semantic segmentation and 3D object detection, our proposed method consistently outperforms previous state-of-the-art models with lower computation cost and measured latency (on a single GTX1080Ti GPU).

\subsection{3D Scene Segmentation}

\begin{table}[t]
\setlength{\tabcolsep}{8pt}
\small\centering
\begin{tabular}{lcccc}
    \toprule
     & \#Params (M) & \#MACs (G) & Latency (ms) & mIoU \\
    \midrule
    PointNet~\cite{qi2017pointnet} & 3.0$^{*}$ & -- & 500$^{*}$ & 14.6 \\
    SPGraph~\cite{landrieu2018large} & 0.3$^{*}$ & -- & 5200$^{*}$ & 17.4 \\
    PointNet++~\cite{qi2017pointnet++} & 6.0$^{*}$ & -- & 5900$^{*}$ & 20.1 \\
    PVCNN~\cite{liu2019point} & 2.5 & 42.4 & 146 & 39.0 \\
    TangentConv~\cite{tatarchenko2018tangent} & 0.4$^{*}$ & -- & 3000$^{*}$ & 40.9 \\
    RandLA-Net~\cite{hu2019randla} & 1.2 & 66.5 & 880 (\textcolor{red}{256}+\textcolor{blue}{624}) & 53.9 \\
    KPConv~\cite{thomas2019kpconv} & 18.3 & 207.3 & -- & 58.8 \\
    MinkowskiNet~\cite{choy20194d} & 21.7 & 114.0 & 294 & 63.1 \\
    \midrule
    \multirow{2}{*}{\textbf{\modelshort} (Ours)} & 2.6 & 15.0 & 110 & 63.7 \\
     & 12.5 & 73.8 & 259 & \textbf{66.4} \\
    \bottomrule
\end{tabular}
\caption{Results of outdoor scene segmentation on SemanticKITTI. \modelshort outperforms MinkowskiNet with \textbf{2.7}$\times$ measured speedup. Here, \textcolor{red}{red numbers} correspond to the computation time, and \textcolor{blue}{blue numbers} correspond to the post-processing time. $^{*}$: results directly taken from Behley~\etal~\cite{behley2019semantickitti}.}
\label{tab:semantickitti:results:3d}
\end{table}

\begin{table}[t]
\setlength{\tabcolsep}{7.5pt}
\small\centering
\begin{tabular}{lcccc}
    \toprule
     & \#Params (M) & \#MACs (G) & Latency (ms) & mIoU \\
    \midrule
    DarkNet21Seg~\cite{behley2019semantickitti} & 24.7 & 212.6 & 73 (\textcolor{red}{49}+\textcolor{blue}{24}) & 47.4 \\
    DarkNet53Seg~\cite{behley2019semantickitti} & 50.4 & 376.3 & 102 (\textcolor{red}{78}+\textcolor{blue}{24}) & 49.9 \\
    SqueezeSegV3-21~\cite{xu2020squeezesegv3} & 9.4 & 187.5 & 97 (\textcolor{red}{73}+\textcolor{blue}{24}) & 51.6 \\
    SqueezeSegV3-53~\cite{xu2020squeezesegv3} & 26.2 & 515.2 & 238 (\textcolor{red}{214}+\textcolor{blue}{24}) & 55.9 \\
    3D-MiniNet~\cite{alonso20203d} & 4.0 & -- & -- & 55.8 \\
    PolarNet~\cite{zhang2020polarnet} & 13.6 & 135.0 & 62 & 57.2 \\
    SalsaNext~\cite{cortinhal2020salsanext} & 6.7 & 62.8 & 71 (\textcolor{red}{47}+\textcolor{blue}{24}) & 59.5 \\
    \midrule
    \textbf{\modelshort} (Ours) & \textbf{1.1} & \textbf{8.9} & 89 & \textbf{60.3} \\
    \bottomrule
\end{tabular}
\caption{Results of outdoor scene segmentation on SemanticKITTI. \modelshort outperforms the 2D projection-based methods with at least \textbf{3.6$\times$} model size reduction and \textbf{7.1$\times$} computation reduction. Here, \textcolor{red}{red numbers} correspond to the computation time, and \textcolor{blue}{blue numbers} correspond to the projection time.}
\label{tab:semantickitti:results:2d}
\end{table}

We first evaluate our method on 3D semantic segmentation and conduct experiments on the large-scale outdoor scene dataset, SemanticKITTI~\cite{behley2019semantickitti}. This dataset contains 23,201 LiDAR point clouds for training and 20,351 for testing, and it is annotated from all 22 sequences in the KITTI~\cite{geiger2013vision} Odometry benchmark. We train all models on the entire training set and report the mean intersection-over-union (mIoU) on the official test set under the single scan setting. We provide additional experimental results (on both validation and test set) in the appendix.

\paragraph{Results.}

\begin{figure*}[t]
\begin{subfigure}[t]{0.325\linewidth}
    \centering
    \includegraphics[width=\linewidth]{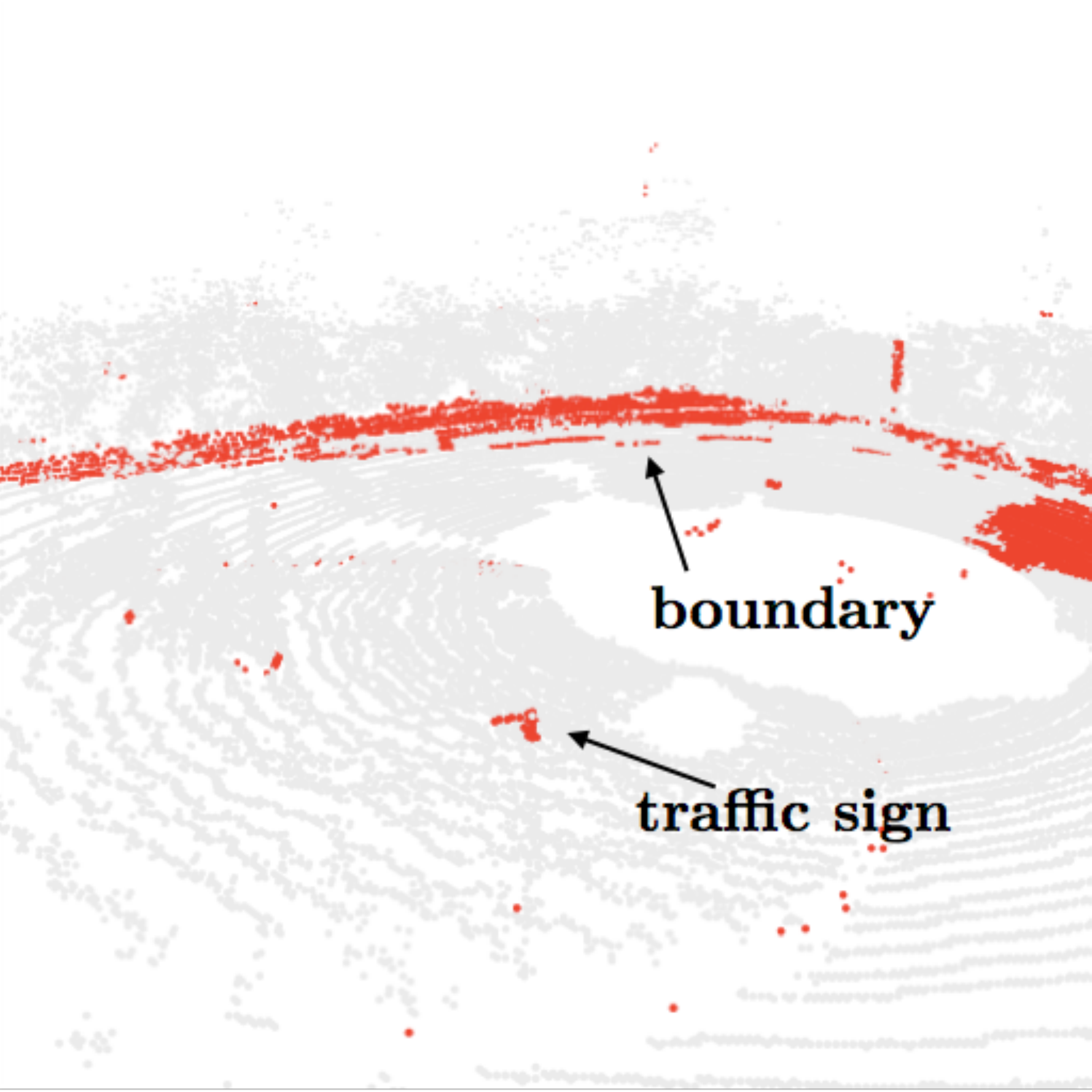}
    \includegraphics[width=\linewidth]{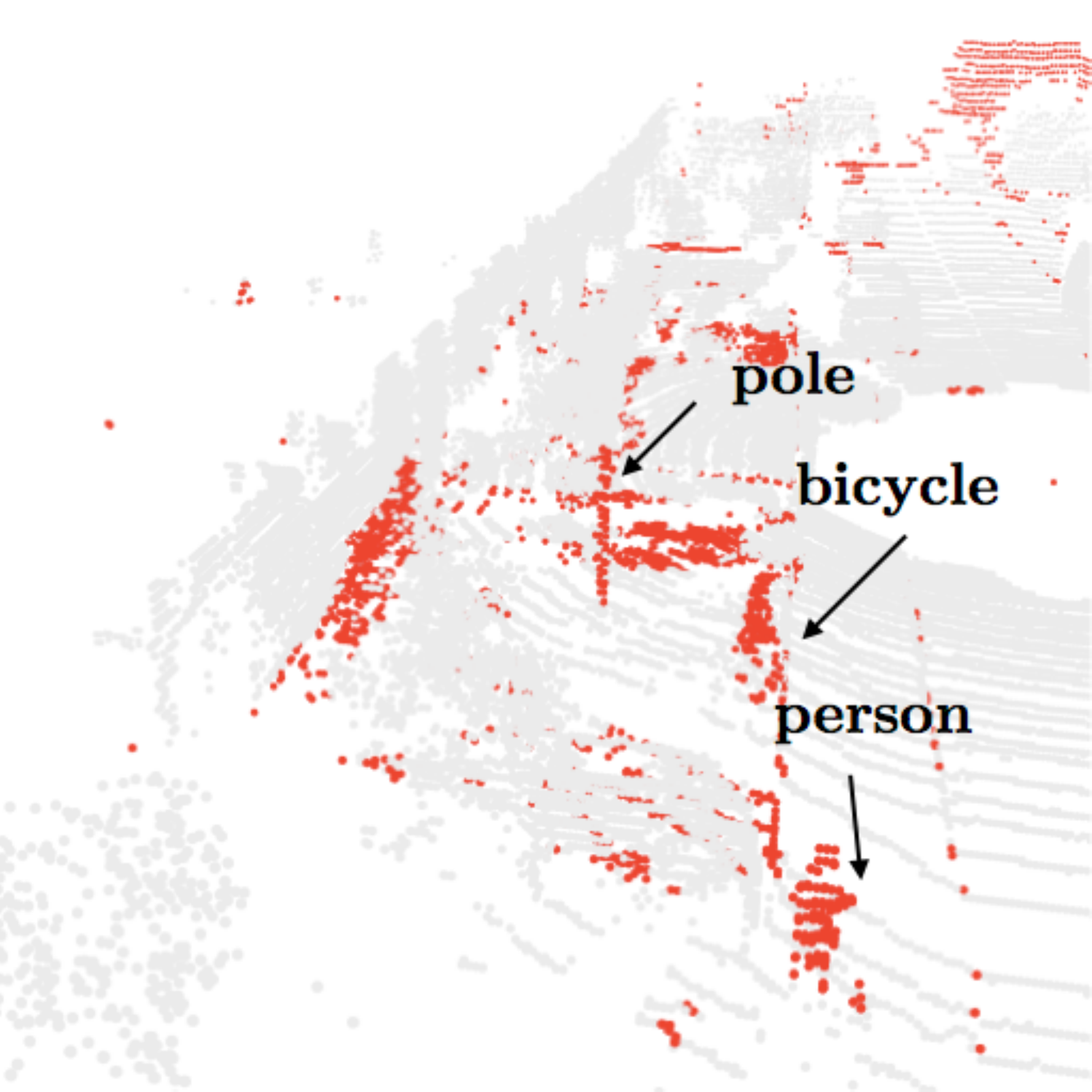}
    \caption{Error by MinkowskiNet}
\end{subfigure}
\begin{subfigure}[t]{0.325\linewidth}
    \centering
    \includegraphics[width=\linewidth]{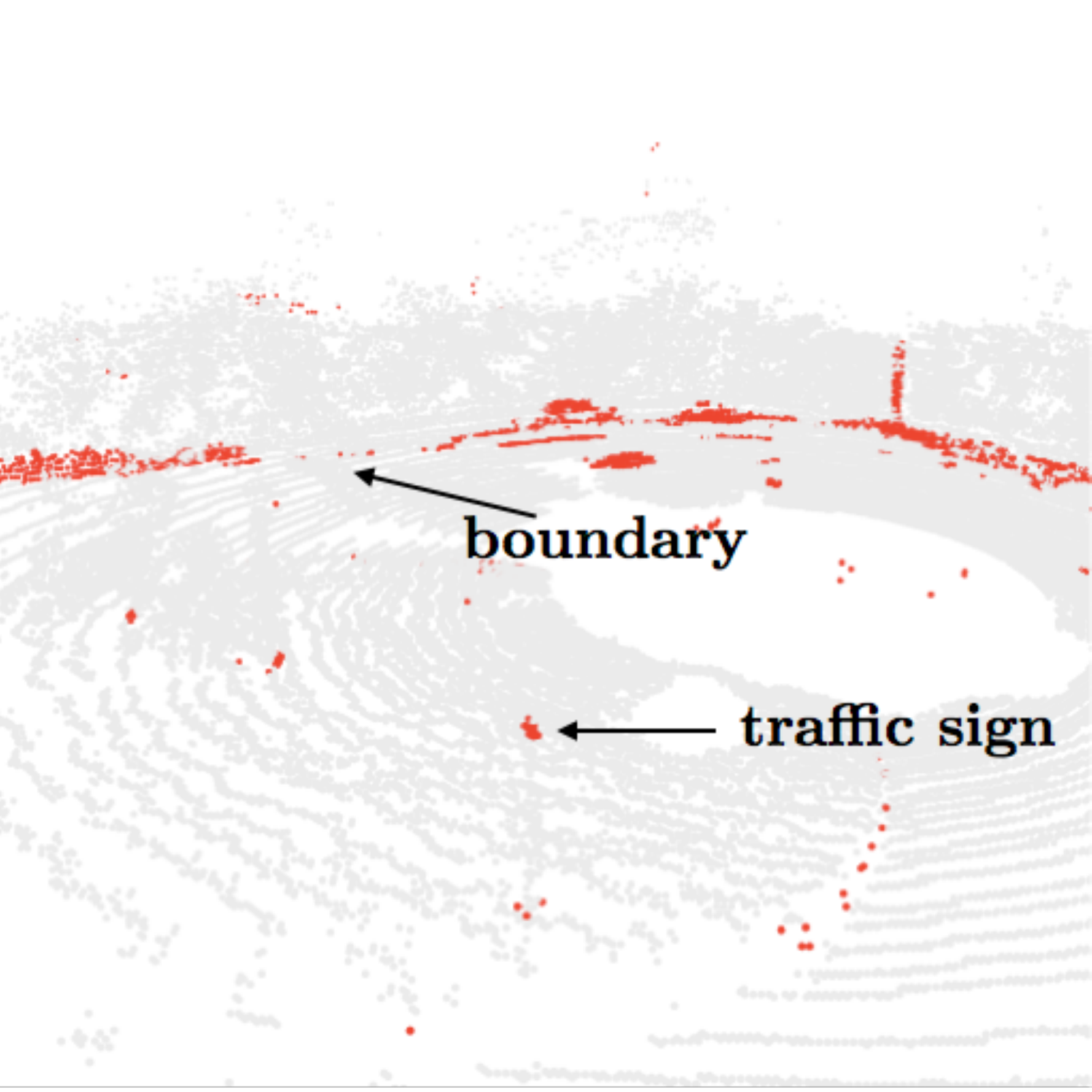}
    \includegraphics[width=\linewidth]{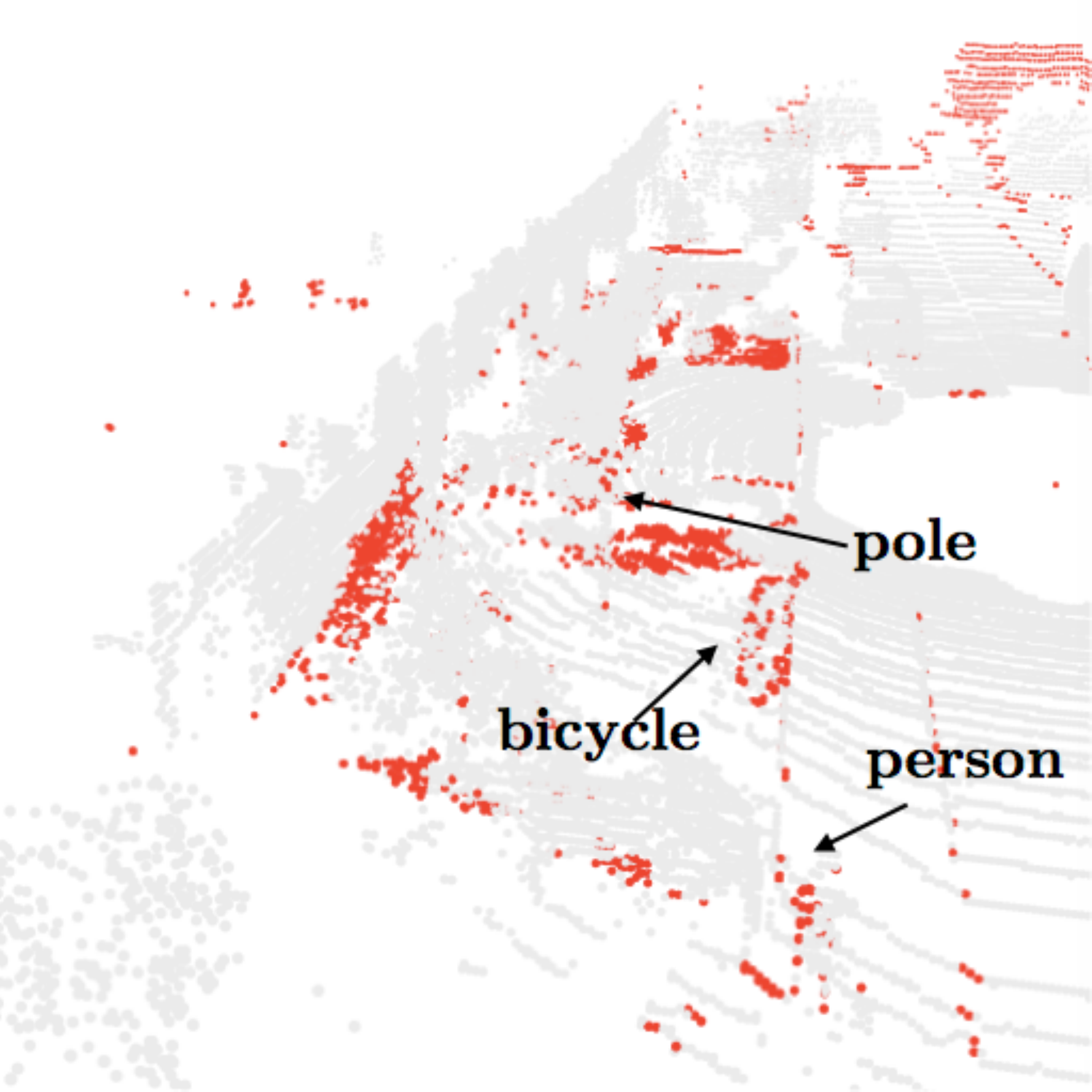}
    \caption{Error by \modelshort}
\end{subfigure}
\begin{subfigure}[t]{0.325\linewidth}
    \centering
    \includegraphics[width=\linewidth]{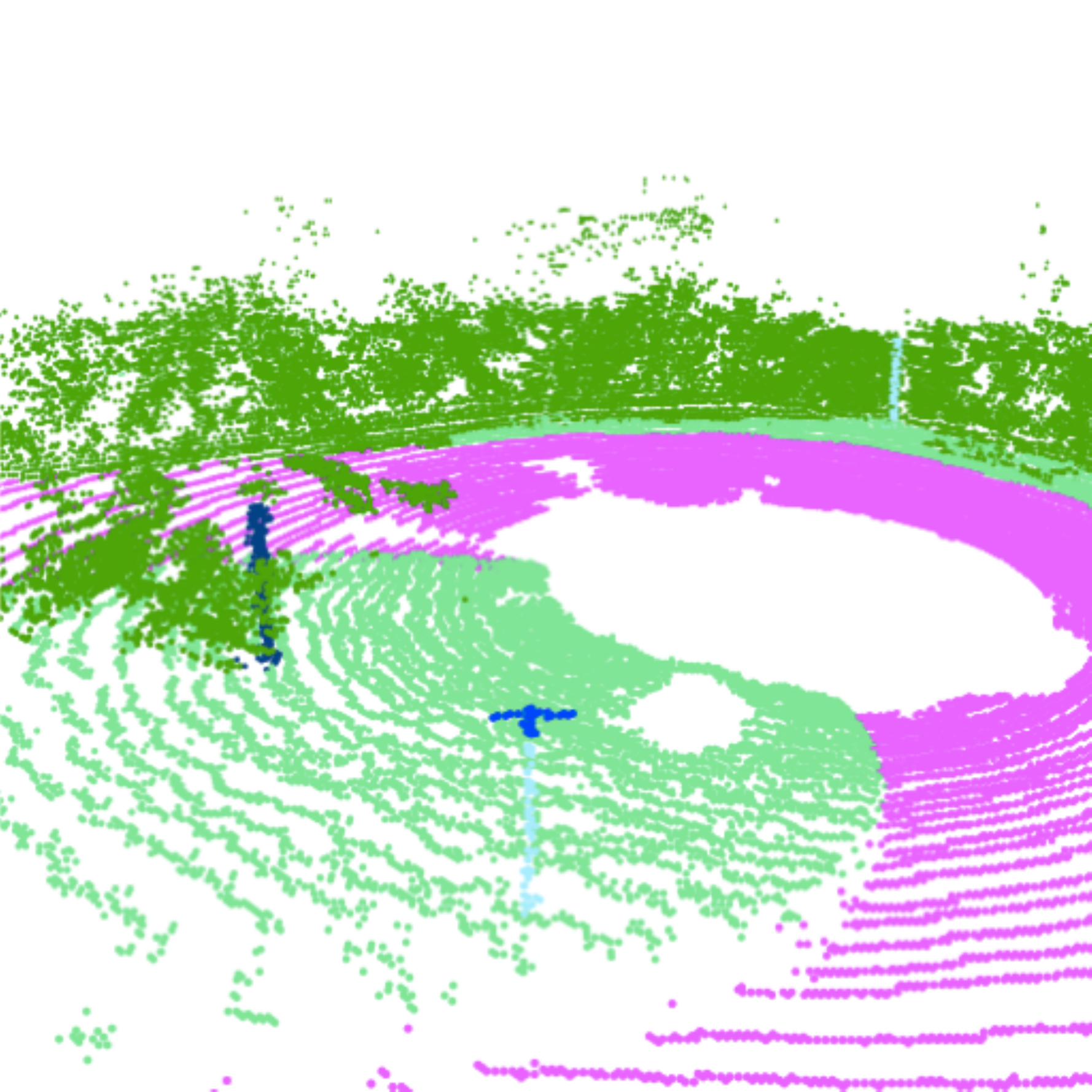}
    \includegraphics[width=\linewidth]{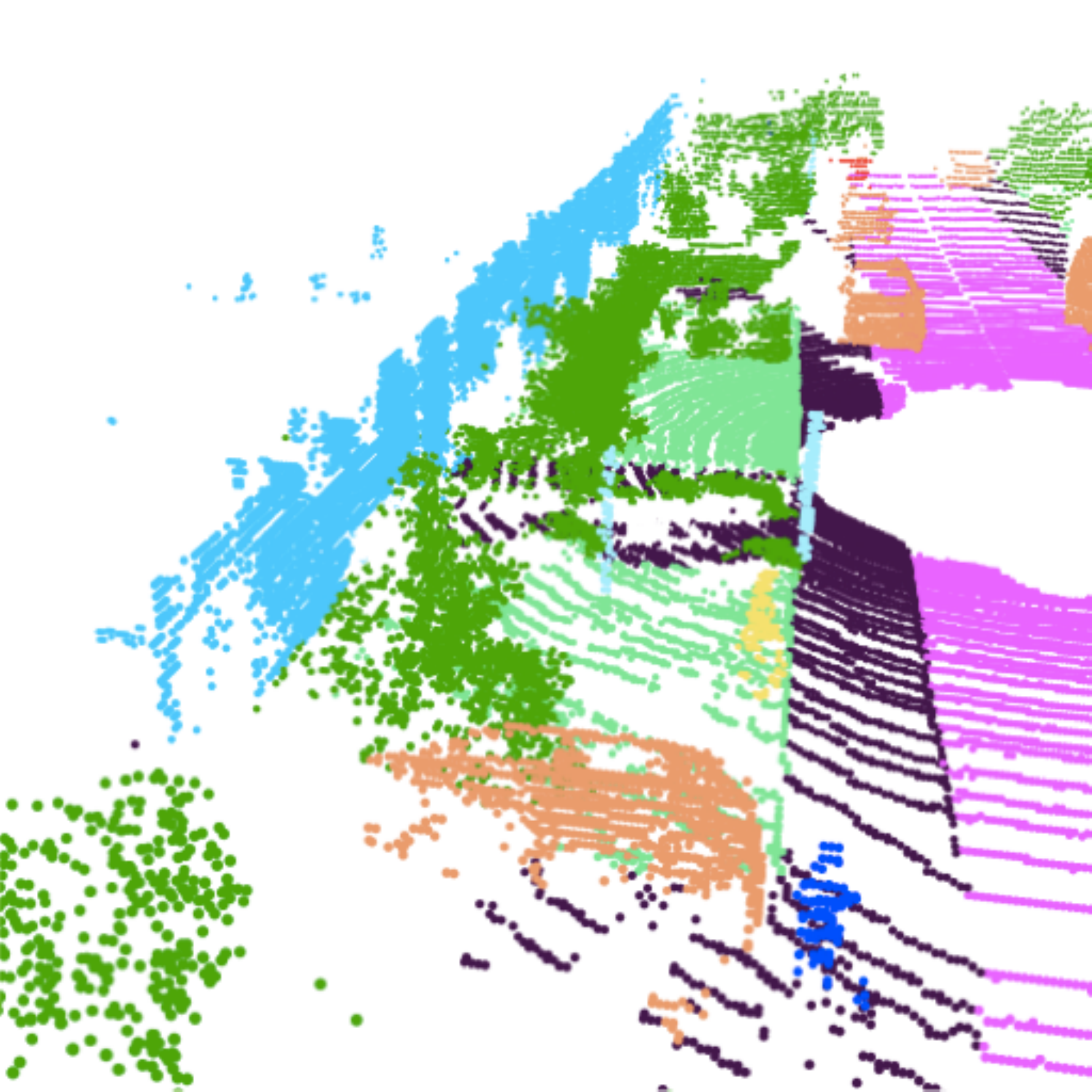}
    \caption{Ground Truth}
\end{subfigure}
\caption{MinkowskiNet has a higher error recognizing small objects and region boundaries, while SPVNAS recognizes small objects better thanks to the high-resolution point-based branch.}
\label{fig:semantickitti:visualizations}
\end{figure*}

As shown in \tab{tab:semantickitti:results:3d}, \modelshort outperforms the previous state-of-the-art MinkowskiNet~\cite{choy20194d} by \textbf{3.3\%} in mIoU with 1.7$\times$ model size reduction, 1.5$\times$ computation reduction and 1.1$\times$ measured speedup. Further, we downscale our \modelshort by setting the resource constraint to 15G MACs. This offers us with a much smaller model that outperforms MinkowskiNet by 0.6\% in mIoU with \textbf{8.3$\times$} model size reduction, \textbf{7.6$\times$} computation reduction, and \textbf{2.7$\times$} measured speedup. In \fig{fig:semantickitti:visualizations}, we also provide some qualitative comparisons between \modelshort and MinkowskiNet: our \modelshort has lower errors especially for small instances.

We further compare our \modelshort with 2D projection-based models in~\tab{tab:semantickitti:results:2d}. With the smaller backbone (by removing the decoder layers), \modelshort outperforms DarkNets~\cite{behley2019semantickitti} by more than \textbf{10\%} in mIoU with 1.2$\times$ measured speedup even though 2D convolutions are much better optimized by modern deep learning libraries. Compared with other 2D methods, \modelshort achieves at least \textbf{8.5}$\times$ model size reduction and \textbf{15.2}$\times$ computation reduction while being much more accurate. Furthermore, our \modelshort achieves higher mIoU than KPConv~\cite{thomas2019kpconv}, which is the previous state-of-the-art point-based model, with \textbf{17$\times$} model size reduction and \textbf{23$\times$} computation reduction.

\begin{table}[t]
\setlength{\tabcolsep}{4.5pt}
\small\centering
\begin{tabular}{lccccccccc}
    \toprule
     & \multicolumn{3}{c}{Car} & \multicolumn{3}{c}{Cyclist} & \multicolumn{3}{c}{Pedestrian} \\
     \cmidrule(lr){2-4}\cmidrule(lr){5-7}\cmidrule(lr){8-10}
     & Easy & Mod. & Hard & Easy & Mod. & Hard & Easy & Mod. & Hard \\
    \midrule
    SECOND~\cite{yan2018second} & 84.7 & 76.0 & 68.7 & 75.8 & 60.8 & 53.7 & 45.3 & 35.5 & 33.1 \\
    SECOND (Repro.) & 87.5 & 77.9 & 74.4 & 76.0 & 59.7 & 52.9 & 49.1 & \textbf{41.7} & \textbf{39.1} \\
    \midrule
    \textbf{\cnnshort} (Ours) & \textbf{87.8} & \textbf{78.4} & \textbf{74.8} & \textbf{80.1} & \textbf{63.7} & \textbf{56.2} & \textbf{49.2} & 41.4 & 38.4 \\
    \bottomrule
\end{tabular}
\caption{Results of outdoor object detection on KITTI. \cnnshort outperforms SECOND in most categories especially for the cyclist.}
\label{tab:kitti:results}
\end{table}

\subsection{3D Object Detection}

We also evaluate our method on 3D object detection and conduct experiments on the outdoor scene dataset, KITTI~\cite{geiger2013vision}. We follow the generally adopted training-validation split, where 3,712 samples are used for training and 3,769 samples are left for validation. We report the mean average precision (mAP) on the test set with 3D IoU thresholds of 0.7 for car, 0.5 for cyclist and pedestrian. We refer the readers to the appendix for more experimental results on the validation set.

\paragraph{Results.}

We compare our method against SECOND~\cite{yan2018second}, the state-of-the-art single-stage model for 3D object detection. SECOND consists of a sparse encoder using 3D Sparse Convolutions and a region proposal network that performs 2D convolutions after projecting the encoded features to the bird's-eye view (BEV). We reimplement and retrain SECOND: our implementation already outperforms the results in the original paper~\cite{yan2018second}. As for our model, we only replace these 3D Sparse Convolutions in SECOND with our SPVConv while keeping all the other settings the same for fair comparison. As summarized in~\tab{tab:kitti:results}, our \cnnshort achieves significant improvement in cyclist detection, for which we argue that the high-resolution point-based branch carries more information for small instances.
\section{Analysis}

Our \modelshort achieves higher accuracy and better efficiency than the previous state-of-the-art MinkowskiNet. In this section, we provide more detailed analysis to better understand the contributions of \moduleshort and \nasshort.

\subsection{\module (\moduleshort)}

\begin{table}[t]
\setlength{\tabcolsep}{7pt}
\small\centering
\begin{tabular}{lccccc}
    \toprule
     & Person & Bicycle & Bicyclist & Motorcycle & Motorcyclist \\
    \midrule
    MinkowskiNet~\cite{choy20194d} & 60.9 & 40.4 & 61.9 & 47.4 & 18.7 \\
    \midrule
    \multirow{2}{*}{\textbf{\modelshort} (Ours)} & 65.7 & 51.6 & 65.2 & 50.8 & 43.7 \\
    & (\textbf{+4.8}) & (\textbf{+11.2}) & (\textbf{+3.3}) & (\textbf{+3.4}) & (\textbf{+25.0}) \\
    \bottomrule
\end{tabular}
\caption{Results of per-class performance on SemanticKITTI. \modelshort has a large advantage on small objects, such as bicyclist and motorcyclist.}
\label{tab:analysis:classes}
\end{table}
\begin{figure*}[t]
\includegraphics[width=\linewidth]{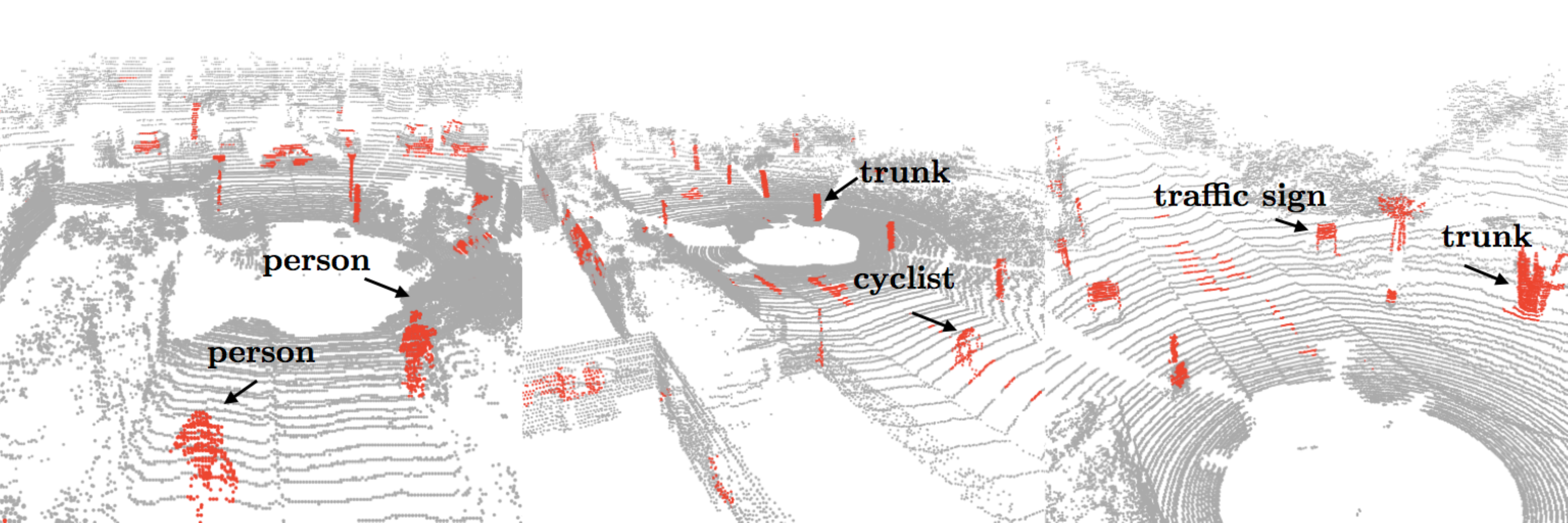}
\caption{The point-based branch learns to put its attention on small instances (\ie, pedestrians, cyclists, traffic signs). Here, the points in red are the ones with the top 5\% largest feature norm in the point-based branch.}
\label{fig:analysis:attention}
\end{figure*}
\begin{figure*}[t]
\includegraphics[width=\linewidth]{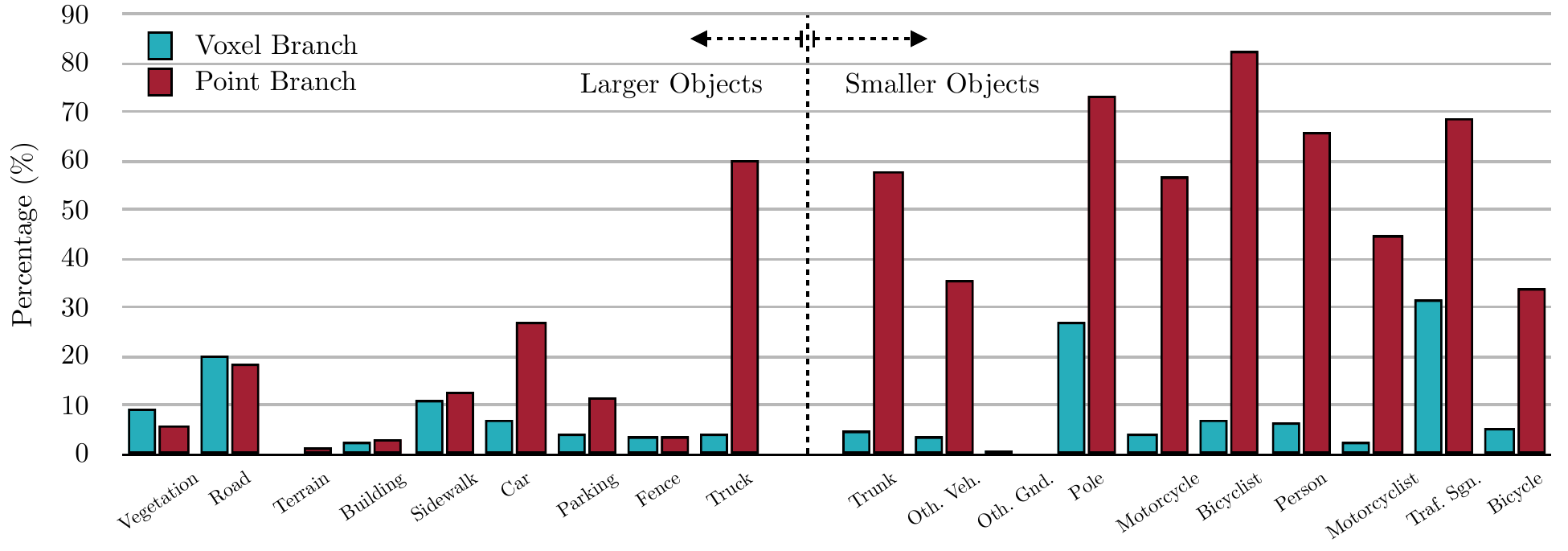}
\caption{Average percentage of activated points in point-based and sparse voxel-based branches for all 19 classes in SemanticKITTI~\cite{behley2019semantickitti}: the point-based branch attends to smaller objects as the red bars are much higher.}
\label{fig:analysis:activation}
\end{figure*}

From \tab{tab:analysis:classes}, our \modelshort has a very large advantage (up to 25\%) on relatively small objects such as pedestrians and cyclists. To explain this, we train \cnnshort on SemanticKITTI with the sequence 08 left out for visualization. In \fig{fig:analysis:attention}, we highlight the points with top 5\% feature norm within the point-based branch (in the final \moduleshort). Clearly, the point-based branch learns to attend to small instances such as pedestrians, cyclists, trunks and traffic signs, which echos with our superior performance on these classes.

Further, we quantitatively analyze the feature norms from both point-based and sparse voxel-based branches. Specifically, we first rank the points from both branches separately based on their feature norms, and then, we mark these points with top 10\% feature norm in each branch as activated. From \fig{fig:analysis:activation}, there are significantly more points in the point-based branch being activated for small instances: \eg, more than 80\% for bicyclist. This suggests that our observation in \fig{fig:analysis:attention} generally holds.

\subsection{\nas (\nasshort)}

\begin{figure*}[t]
\begin{subfigure}[t]{0.49\linewidth}
    \centering
    \includegraphics[width=\linewidth]{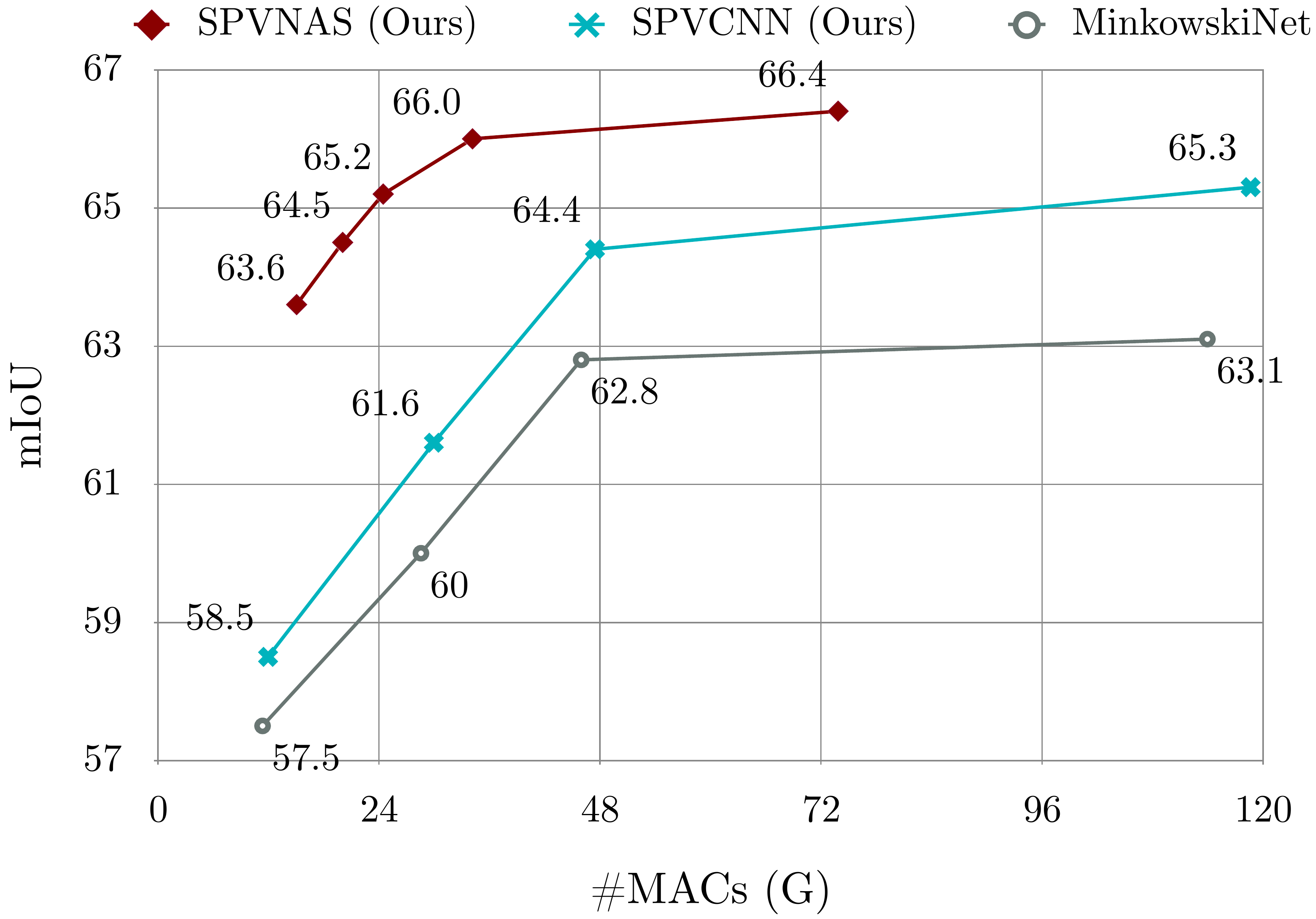}
    \caption{Trade-off: Mean IoU \vs \#MACs}
\end{subfigure}
\begin{subfigure}[t]{0.495\linewidth}
    \centering
    \includegraphics[width=\linewidth]{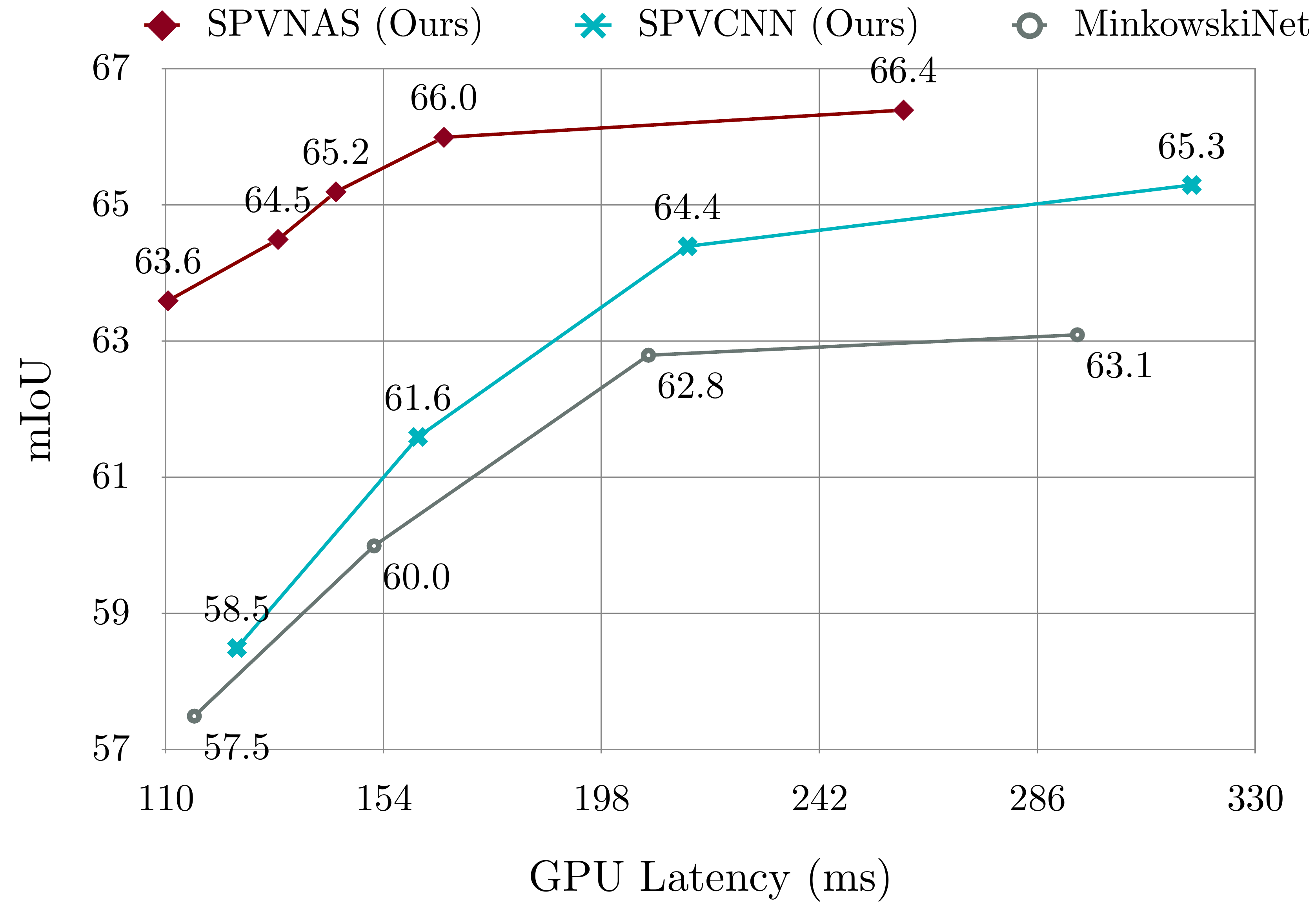}
    \caption{Trade-off: Mean IoU \vs GPU Latency}
\end{subfigure}
\caption{An efficient 3D module (\moduleshort) and a well-designed network architecture (\nasshort) are equally important to the final performance of \modelshort: \textbf{7.6}$\times$ computation reduction and \textbf{2.7}$\times$ measured speedup over MinkowskiNet.}
\label{fig:semantickitti:tradeoffs}
\end{figure*}
\begin{figure*}[t]
\centering
\begin{subfigure}[t]{0.49\linewidth}
    \centering
    \includegraphics[width=\linewidth]{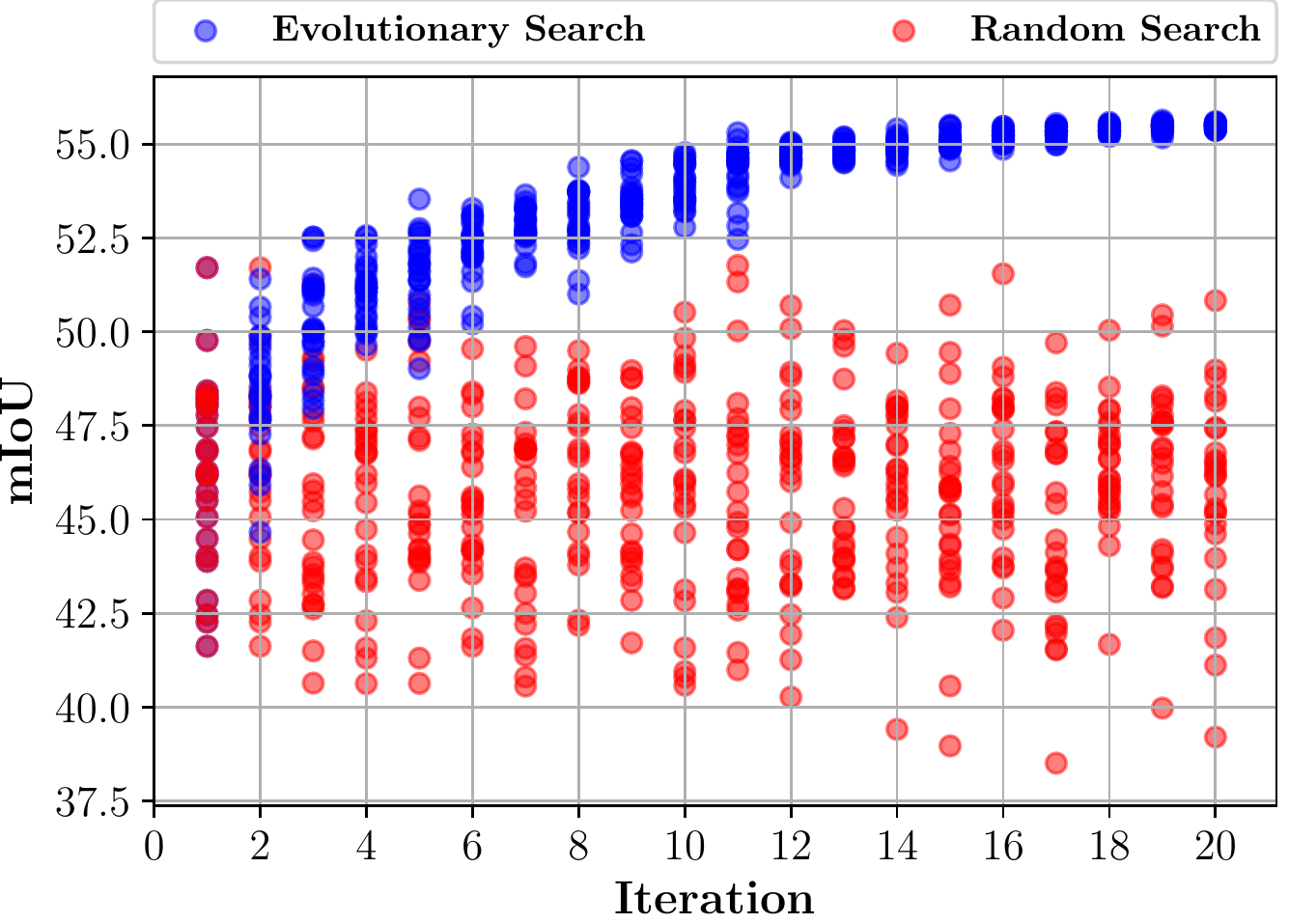}
    \caption{Search curves of ES and RS.}
    \label{fig:analysis:search:a}
\end{subfigure}
\begin{subfigure}[t]{0.49\linewidth}
    \centering
    \includegraphics[width=\linewidth]{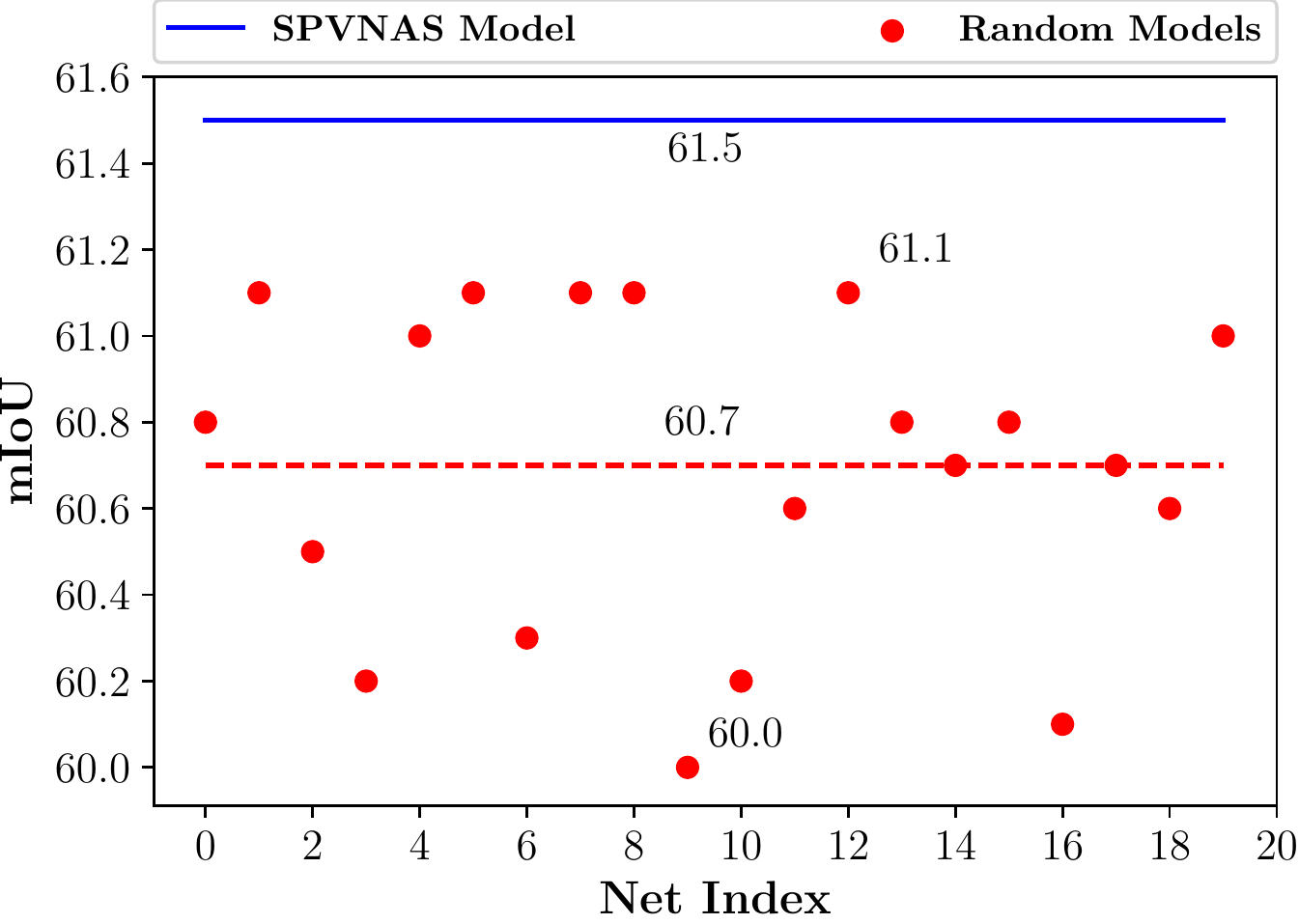}
    \caption{Comparison with random models.}
    \label{fig:analysis:search:b}
\end{subfigure}
\caption{Evolutionary Search (ES) is more sample-efficient than Random Search (RS).}
\label{fig:analysis:search}
\end{figure*}

In \fig{fig:semantickitti:tradeoffs}, we present both mIoU \vs \#MACs and mIoU \vs latency trade-offs, where we uniformly scale the channel numbers in MinkowskiNet and \cnnshort down as our baselines. It can be observed that a better 3D module (\moduleshort) and a well-designed network architecture (\nasshort) are equally important to the final performance boost. Remarkably, \modelshort outperforms MinkowskiNet by more than 6\% in mIoU at 110 ms latency. Such a large improvement comes from the non-uniform channel scaling and elastic nework depth. In these manually-designed models (MinkowskiNet and \cnnshort), 77\% of the total computation is distributed to the upsampling stage. With \nasshort, this ratio is reduced to 47-63\%, making the computation more balanced and the downsampling stage (\ie, feature extraction) more emphasized.

We also compare our evolutionary search with the random architecture search to show that the success of \nasshort does not entirely come from the search space. As in \fig{fig:analysis:search:a}, random architecture search has poor sample efficiency: the best model at the 20\textsuperscript{th} generation performs even worse than the best model in the first generation. In contrast, our evolutionary search is capable of progressively finding better architecture, and the final best architecture performs around 3\% better than the one in the first generation. We also retrain 20 random models sampled from our search space and compare them with \modelshort in \fig{fig:analysis:search:b}. As a result, our \modelshort performs 0.8\% better compared with the average performance of these random models.

\section{Conclusion}

We present \module (\moduleshort), a lightweight 3D module specialized for small object recognition. With \cnnshort built upon \moduleshort, we solve the problem that Sparse Convolution cannot always keep high-resolution representation and Point-Voxel Convolution does not scale up to large 3D scenes. Furthermore, we introduce \nasshort, the first architecture search framework for 3D scene understanding that greatly improves the efficiency and performance of \cnnshort. Extensive experiments on outdoor 3D scene benchmarks demonstrate that the resulting \modelshort model is lightweight, fast and powerful. We hope that this work will inspire future research on efficient 3D deep learning.
\paragraph{Acknowledgements.}

We thank Nick Stathas and Yue Dong for their feedback on the draft. This work is supported by MIT Quest for Intelligence, MIT-IBM Watson AI Lab, Xilinx and Samsung. We thank AWS Machine Learning Research Awards for providing the computational resource.

\bibliographystyle{splncs04}
\bibliography{reference}

\clearpage
\appendix
\renewcommand{\thesection}{A.\arabic{section}}
\renewcommand{\thefigure}{A\arabic{figure}}
\setcounter{section}{0}
\setcounter{figure}{0}

\section{Implementation Details}

We provide more implementation details on how to build our backbone network (\cnnshort), and train the super network and search for the best model (\nasshort).

\subsection{\cnnshort: Backbone Network}

Based on MinkowskiNet~\cite{choy20194d}, we build our backbone network by wrapping residual Sparse Convolution blocks with the high-resolution point-based branch. Specifically, the first \moduleshort voxelizes before the first layer and devoxelizes after the stemming stage (\ie, before the first downsampling). The second \moduleshort voxelizes right after the first \moduleshort and devoxelizes after all four downsampling stages. The final two \moduleshort's each wraps around two upsampling stages.

Also, we design a smaller backbone based on PVCNN~\cite{liu2019point} by directly replacing each volumetric convolution with one convolution layer (followed by normalization and activation layers) and two residual Sparse Convolution blocks.

\subsection{\nasshort: Architecture Search}

We train the super network for 15 epochs that supports the fine-grained channel setting with a starting learning rate 0.24 and cosine learning rate decay. Then, we train for another 15 epochs to incorporate elastic network depth with a starting learning rate 0.096 and cosine learning rate decay. After that, we perform evolutionary architecture search with a population of 50 candidates for 20 generations on the official validation set (sequence 08). Best architecture is directly extracted from the super network and submitted to the test server after finetuning for 10 epochs with a starting learning rate of 0.032 and cosine learning rate decay.

\section{More Results}

\subsection{3D Scene Segmentation}

We present more detailed results of MinkowskiNet~\cite{choy20194d}, \cnnshort and \modelshort on both the official test set and validation set (sequence 08) of SemanticKITTI~\cite{behley2019semantickitti} in \tab{tab:semantickitti:results_test} and \tab{tab:semantickitti:results_val}. For the results on the validation set in \tab{tab:semantickitti:results_val}, we run the whole architecture search pipeline again on sequences 00-07 and 09, leaving sequence 10 out as the mini-validation set, and report the results on sequence 08. We observe similar trends on both test and validation results: both a better 3D module (\moduleshort) and our \nas (\nasshort) pipeline improve the performance of MinkowskiNet~\cite{choy20194d}. 

\begin{table}[t]
\setlength{\tabcolsep}{7pt}
\small\centering
\begin{tabular}{lcccc}
    \toprule
     & \#Params (M) & \#MAdds (G) & Latency (ms) & Mean IoU \\
    \midrule
    MinkowskiNet~\cite{choy20194d} & 2.2 & 11.3 & 115.7 & 57.5 \\
    \textbf{\cnnshort} (Ours) & 2.2 & 11.9 & 124.3 & 58.5 \\
    \textbf{\modelshort} (Ours) & 2.6 & 15.0 & \textbf{110.4} & \textbf{63.6} \\ \midrule
    
    MinkowskiNet~\cite{choy20194d} & 5.5 & 28.5 & 152.0 & 60.0 \\
    \textbf{\cnnshort} (Ours) & 5.5 & 30.0 & 160.9 & 61.6 \\
    \textbf{\modelshort} (Ours) & \textbf{4.2} & \textbf{20.0} & \textbf{132.6} & \textbf{64.5} \\ \midrule
    MinkowskiNet~\cite{choy20194d} & 8.8 & 45.9 & 207.4 & 62.8 \\
    \textbf{\cnnshort} (Ours) & 8.8 & 47.4 & 214.3 & 64.4 \\
    \textbf{\modelshort} (Ours) & \textbf{5.1} & \textbf{24.4} & \textbf{144.3} & \textbf{65.2} \\ \midrule
    MinkowskiNet~\cite{choy20194d} & 21.7 & 113.9 & 294.0 & 63.1 \\
    \textbf{\cnnshort} (Ours) & 21.8 & 118.6 & 317.1 & 63.8 \\
    \textbf{\modelshort} (Ours) & \textbf{7.5} & \textbf{34.1} & \textbf{166.1} & 66.0 \\ 
    \textbf{\modelshort} (Ours) & 12.5 & 73.8 & 259.9 & \textbf{66.4} \\ 
    \bottomrule
\end{tabular}
\caption{Results of 3D scene segmentation on the test set of SemanticKITTI~\cite{behley2019semantickitti}.}
\label{tab:semantickitti:results_test}
\end{table}

\begin{table}[t]
\setlength{\tabcolsep}{7pt}
\small\centering
\begin{tabular}{lcccc}
    \toprule
     & \#Params (M) & \#MAdds (G) & Latency (ms) & Mean IoU \\
    \midrule
    MinkowskiNet~\cite{choy20194d} & 5.5 & 28.5 & 152.0 & 58.9 \\
    \textbf{\cnnshort} (Ours) & 5.5 & 30.0 & 160.9 & 60.7 \\
    \textbf{\modelshort} (Ours) & \textbf{4.5} & \textbf{24.6} & 158.1 & \textbf{62.9} \\ \midrule
    MinkowskiNet~\cite{choy20194d} & 8.8 & 45.9 & 207.4 & 60.3 \\
    \textbf{\cnnshort} (Ours) & 8.8 & 47.4 & 214.3 & 61.4 \\
    \textbf{\modelshort} (Ours) & \textbf{7.0} & \textbf{34.7} & \textbf{175.8} & \textbf{63.5} \\ \midrule
    MinkowskiNet~\cite{choy20194d} & 21.7 & 113.9 & 294.0 & 61.1 \\
    \textbf{\cnnshort} (Ours) & 21.8 & 118.6 & 317.1 & 63.8 \\
    \textbf{\modelshort} (Ours) & \textbf{10.8} & \textbf{64.5} & \textbf{248.7} & \textbf{64.7} \\ 
    \bottomrule
\end{tabular}
\caption{Results of 3D scene segmentation on the validation set of SemanticKITTI~\cite{behley2019semantickitti}.}
\label{tab:semantickitti:results_val}
\end{table}

\subsection{3D Object Detection}

We further provide the results of \cnnshort on the validation set of KITTI~\cite{geiger2012kitti} in \tab{tab:kitti:results_val}. We train both SECOND~\cite{yan2018second} and our \cnnshort on the training set for three times, and we report the average results to reduce the variance. Similar to the results on the test set, \cnnshort also has consistent improvement in almost all classes on the validation set.

\begin{table}[t]
\setlength{\tabcolsep}{4.5pt}
\small\centering
\begin{tabular}{lccccccccc}
    \toprule
     & \multicolumn{3}{c}{Car} & \multicolumn{3}{c}{Cyclist} & \multicolumn{3}{c}{Pedestrian} \\
     \cmidrule(lr){2-4}\cmidrule(lr){5-7}\cmidrule(lr){8-10}
     & Easy & Mod. & Hard & Easy & Mod. & Hard & Easy & Mod. & Hard \\
    \midrule
    SECOND~\cite{yan2018second} & 89.8 & 80.9 & 78.4 & 82.5 & 62.8 & 58.9 & \textbf{68.3} & 60.8 & 55.3 \\
    \midrule
    \textbf{\cnnshort} (Ours) & \textbf{90.9} & \textbf{81.8} & \textbf{79.2} & \textbf{85.1} & \textbf{63.8} & \textbf{60.1} & 68.2 & \textbf{61.6} & \textbf{55.9} \\
    \bottomrule
\end{tabular}
\caption{Results of 3D object detection on the validation set of KITTI~\cite{geiger2013vision}.}
\label{tab:kitti:results_val}
\end{table}

\section{More Visualizations}

\subsection{3D Scene Segmentation}

\begin{figure*}[ht!]
    \centering
    \begin{subfigure}[t]{0.32\linewidth}
        \centering
        \includegraphics[width=\linewidth]{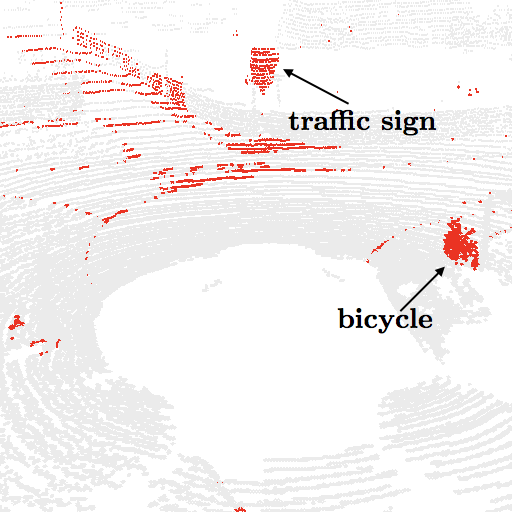}
        \includegraphics[width=\linewidth]{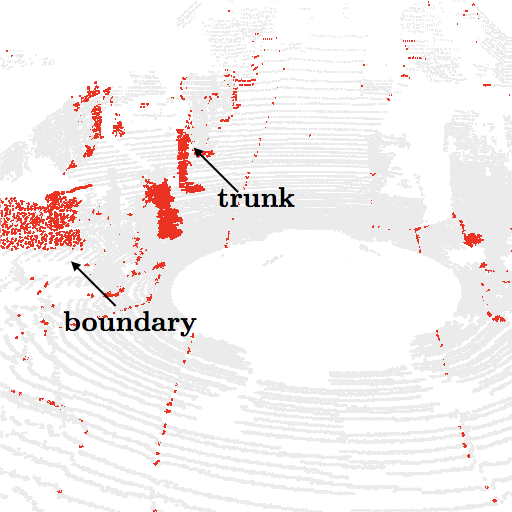}
        \includegraphics[width=\linewidth]{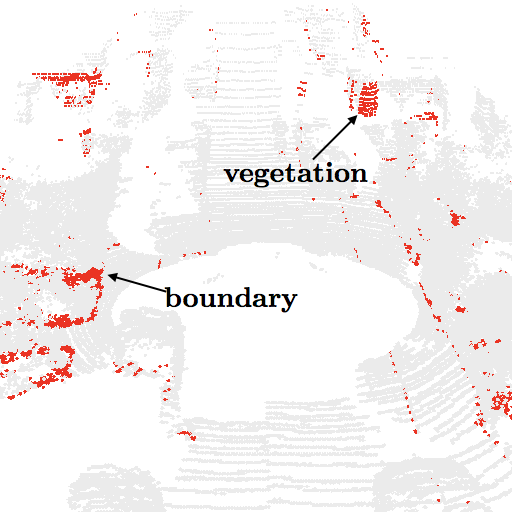}
        \caption{Error by MinkowskiNet}
        \label{fig:semantickitti:appendix:mink}
    \end{subfigure}
    \begin{subfigure}[t]{0.32\linewidth}
        \centering
        \includegraphics[width=\linewidth]{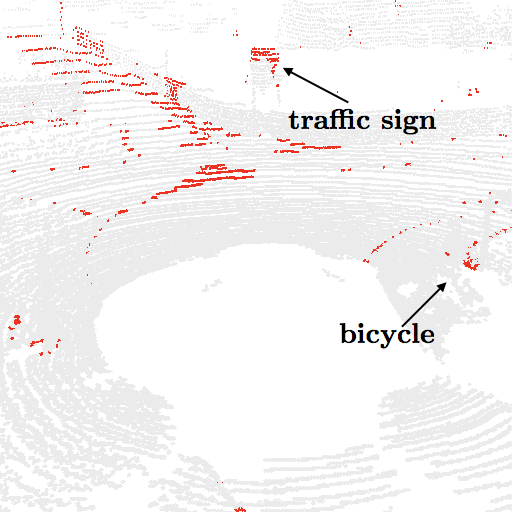}
        \includegraphics[width=\linewidth]{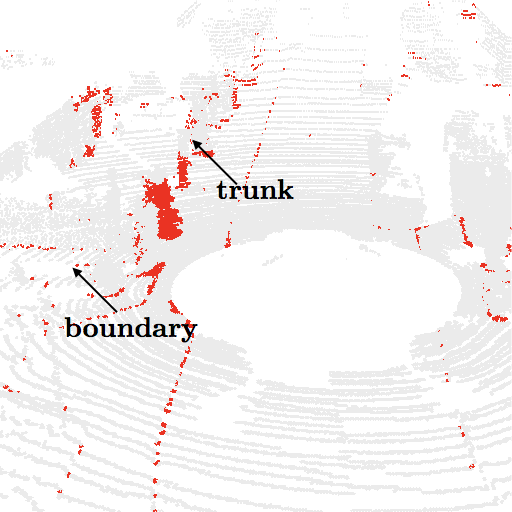}
        \includegraphics[width=\linewidth]{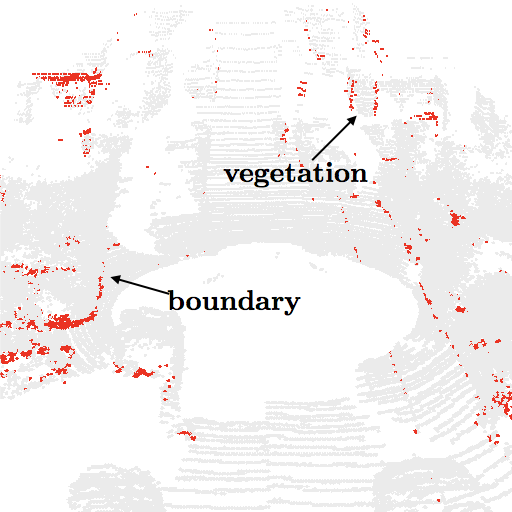}
        \caption{Error by \modelshort}
        \label{fig:semantickitti:appendix:spvnas}
    \end{subfigure}
    \begin{subfigure}[t]{0.32\linewidth}
        \centering
        \includegraphics[width=\linewidth]{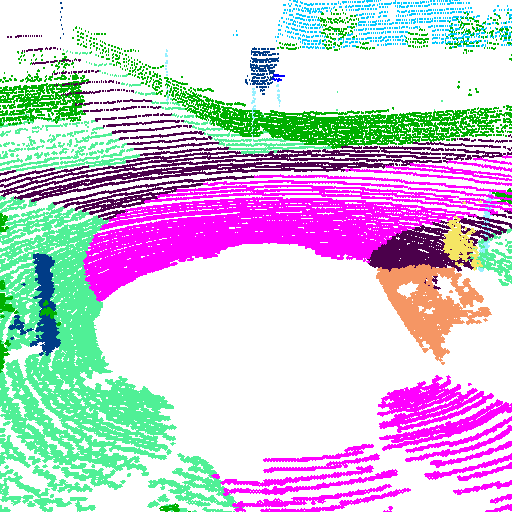}
        \includegraphics[width=\linewidth]{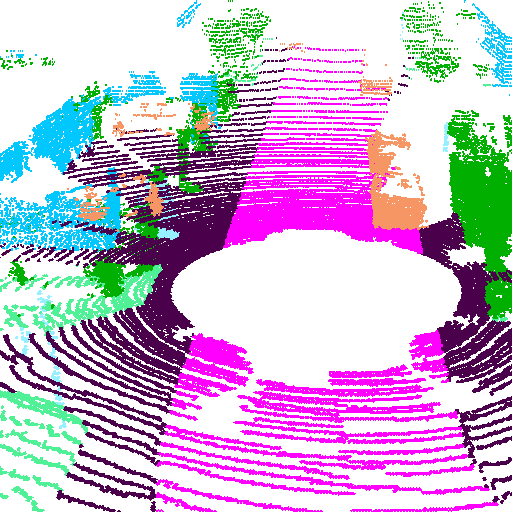}
        \includegraphics[width=\linewidth]{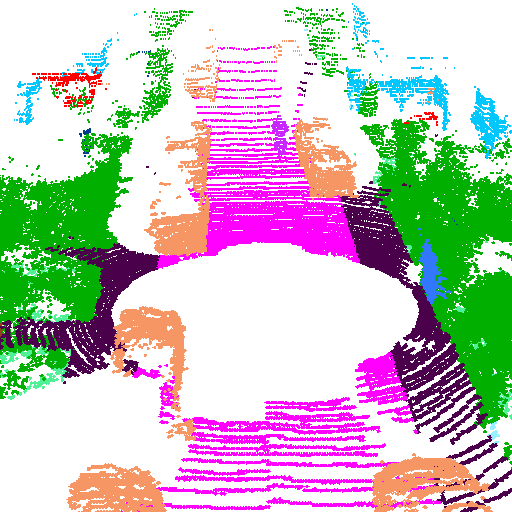}
        \caption{Ground Truth}
        \label{fig:semantickitti:appendix:gt}
    \end{subfigure}
    \caption{Qualitative comparisons between MinkowskiNet~\cite{choy20194d} and \modelshort.}
    \vspace{-8pt}
    \label{fig:semantickitti:appendix:visualizations}
\end{figure*}

We provide more visualizations for MinkowskiNet~\cite{choy20194d} and \modelshort in \fig{fig:semantickitti:appendix:visualizations} to demonstrate that the improvements brought by \moduleshort on small objects and region boundaries are general. For instance, in the first row, MinkowskiNet segments the entire traffic sign and bicycle instances incorrectly; however, our \modelshort is capable of making almost no mistakes on these very small objects. Also, we observe in next two rows that MinkowskiNet does not perform well on the sidewalk-building or sidewalk-vegetation boundaries where our \modelshort has a clear advantage.

\begin{figure*}[ht!]
    \centering
    \begin{subfigure}[t]{0.32\linewidth}
        \centering
        \includegraphics[width=\linewidth]{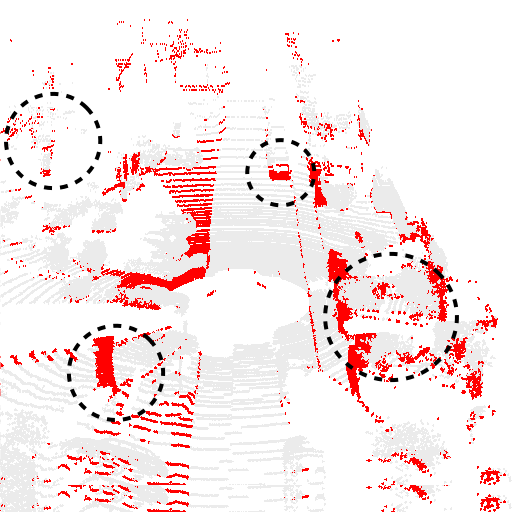}
        \includegraphics[width=\linewidth]{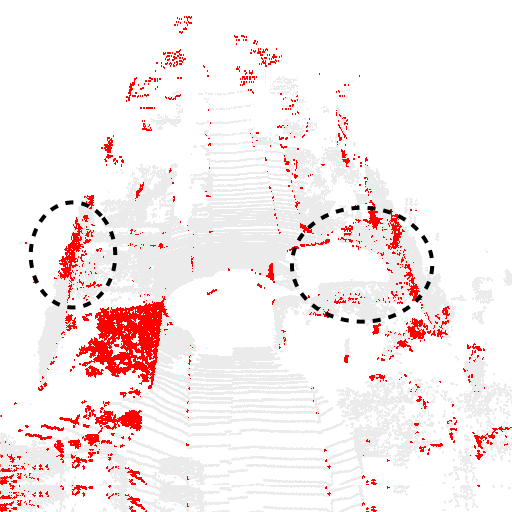}
        \includegraphics[width=\linewidth]{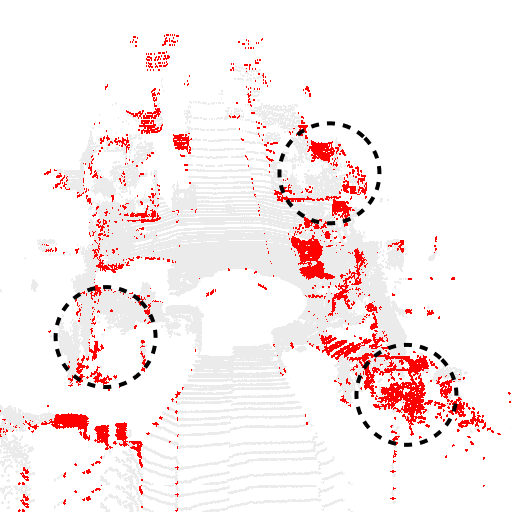}
        \includegraphics[width=\linewidth]{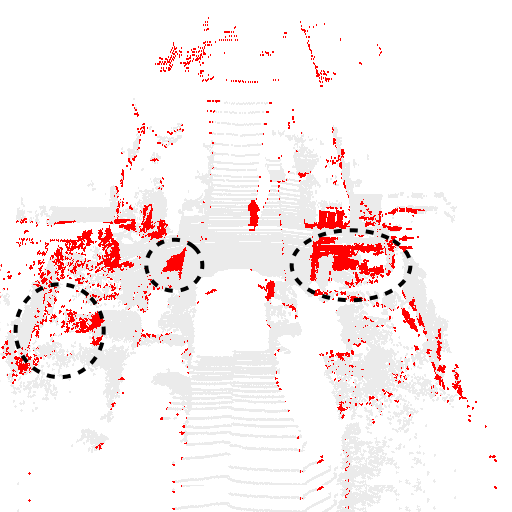}
        \caption{Error by DarkNet}
        \label{fig:semantickitti:appendix:mink:small}
    \end{subfigure}
    \begin{subfigure}[t]{0.32\linewidth}
        \centering
        \includegraphics[width=\linewidth]{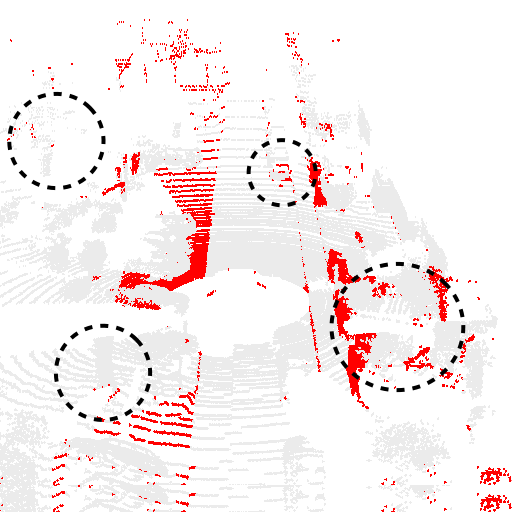}
        \includegraphics[width=\linewidth]{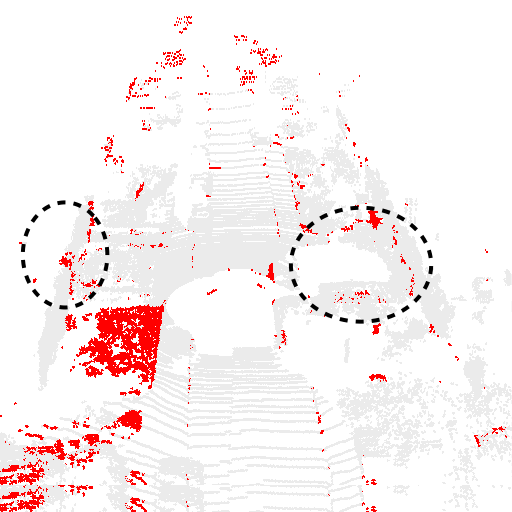}
        \includegraphics[width=\linewidth]{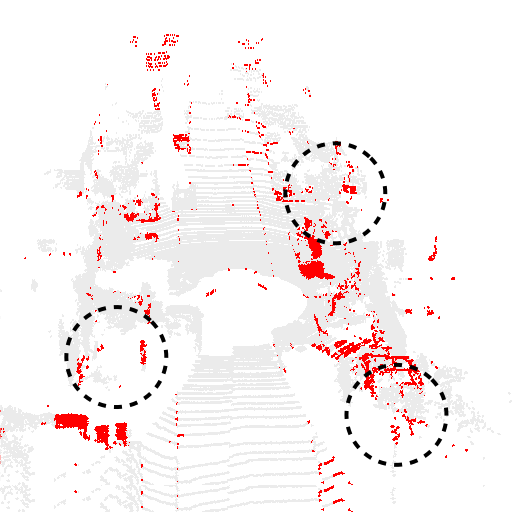}
        \includegraphics[width=\linewidth]{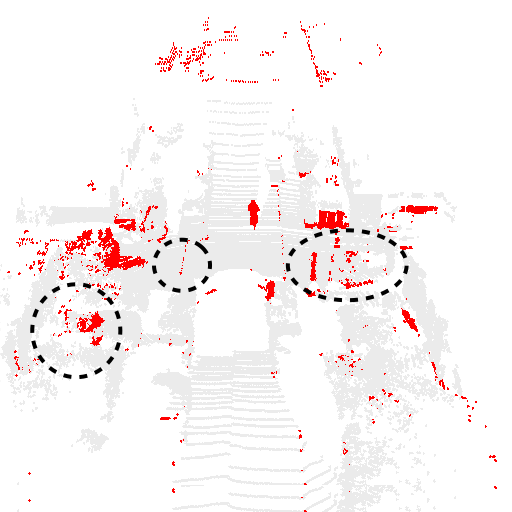}
        \caption{Error by \modelshort}
        \label{fig:semantickitti:appendix:pv:small}
    \end{subfigure}
    \begin{subfigure}[t]{0.32\linewidth}
        \centering
        \includegraphics[width=\linewidth]{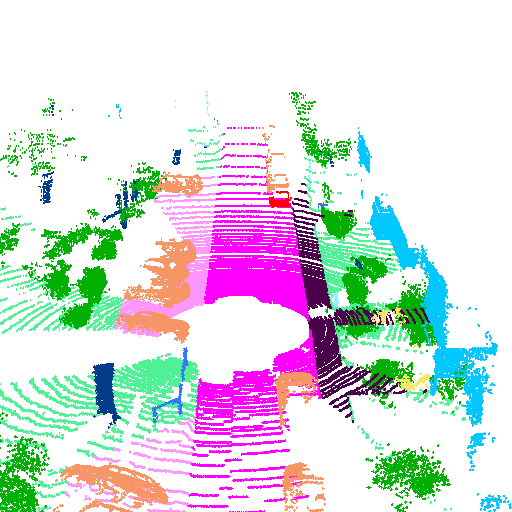}
        \includegraphics[width=\linewidth]{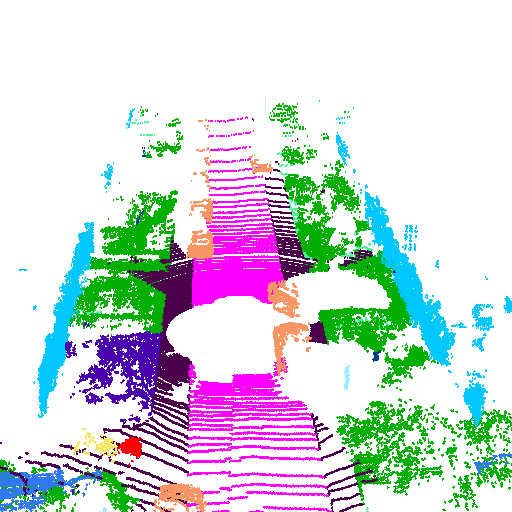}
        \includegraphics[width=\linewidth]{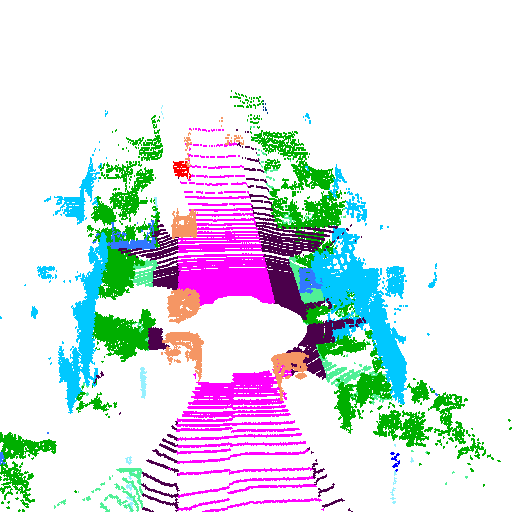}
        \includegraphics[width=\linewidth]{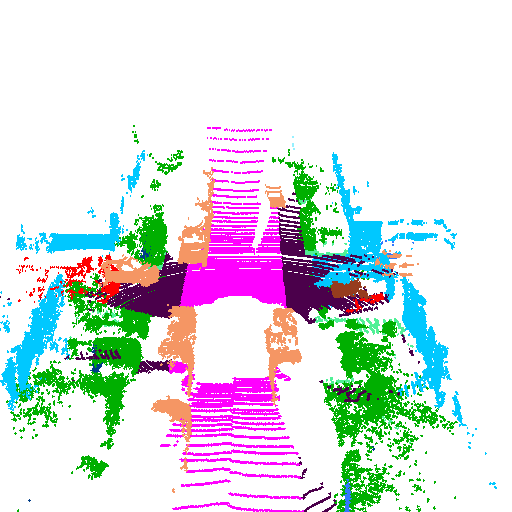}
        \caption{Ground Truth}
        \label{fig:semantickitti:appendix:gt:small}
    \end{subfigure}
    \caption{Qualitative comparisons between DarkNet53~\cite{behley2019semantickitti} and \modelshort. }
    \vspace{-8pt}
    \label{fig:semantickitti:appendix:visualizations:small}
\end{figure*}

In \fig{fig:semantickitti:appendix:visualizations:small}, we compare our smaller \modelshort with DarkNet53~\cite{behley2019semantickitti}. DarkNet53 uses spherical projections to project 3D point clouds to a 2D plane such that part of the geometry information is lost. Our \modelshort, in contrast, directly learns on 3D data and is more powerful in the geometric modeling. We observe that \modelshort has significant advantages in both large regions (last three rows) and smaller instances (first row).

\subsection{3D Object Detection}

\begin{figure*}[ht!]
    \centering
    \begin{subfigure}[t]{0.32\linewidth}
        \centering
        \includegraphics[width=\linewidth]{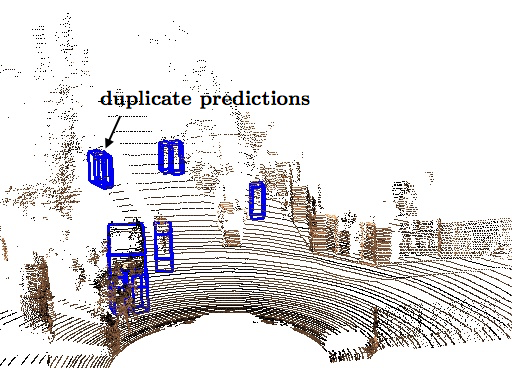}
        \includegraphics[width=\linewidth]{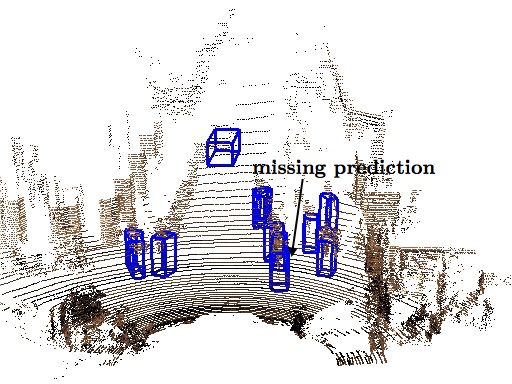}
        \includegraphics[width=\linewidth]{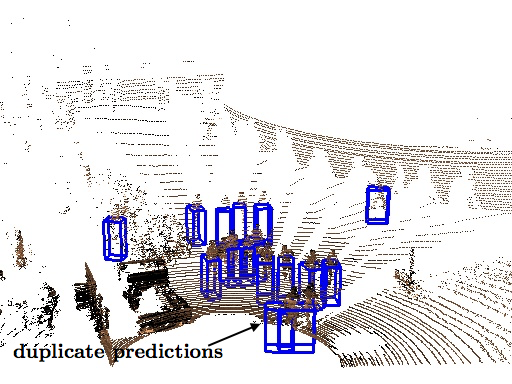}
        \includegraphics[width=\linewidth]{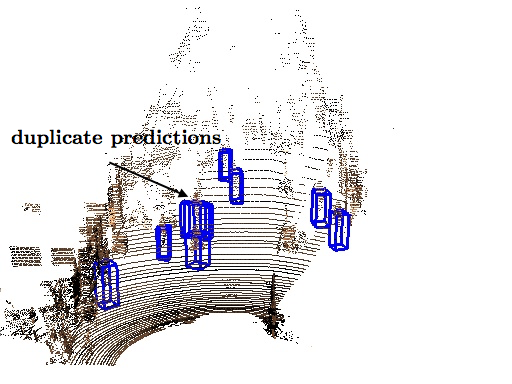}
        \caption{SECOND}
        \label{fig:semantickitti:appendix:mink:det}
    \end{subfigure}
    \begin{subfigure}[t]{0.32\linewidth}
        \centering
        \includegraphics[width=\linewidth]{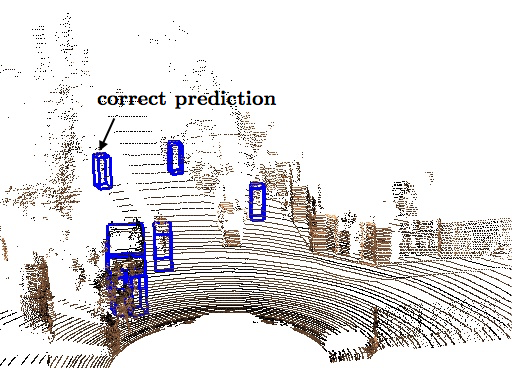}
        \includegraphics[width=\linewidth]{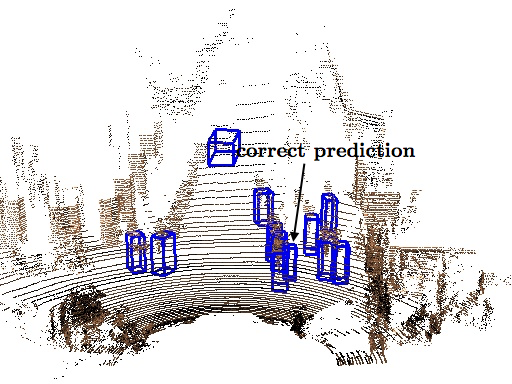}
        \includegraphics[width=\linewidth]{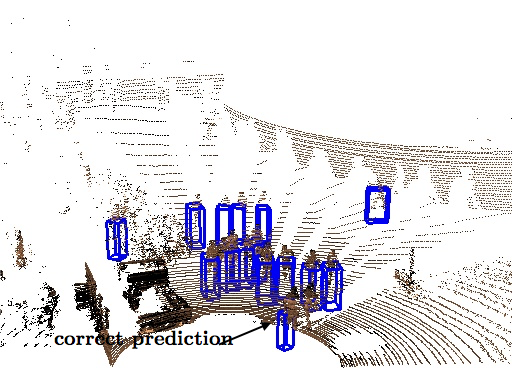}
        \includegraphics[width=\linewidth]{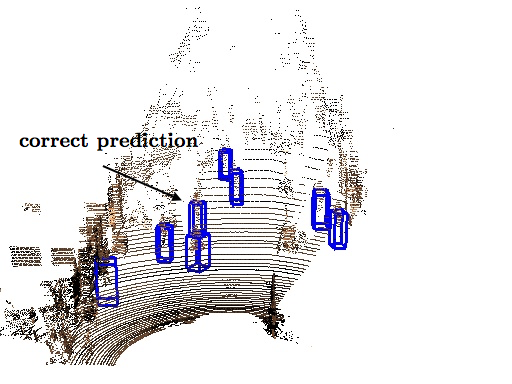}
        \caption{\cnnshort}
        \label{fig:semantickitti:appendix:pv:det}
    \end{subfigure}
    \begin{subfigure}[t]{0.32\linewidth}
        \centering
        \includegraphics[width=\linewidth]{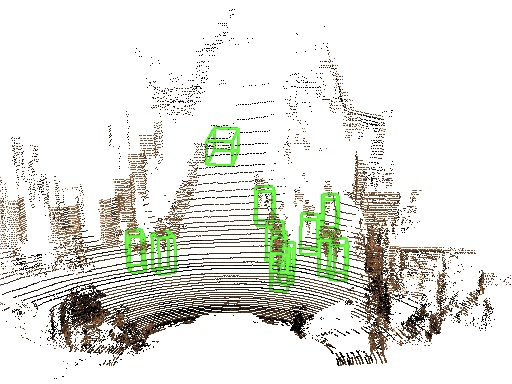}
        \includegraphics[width=\linewidth]{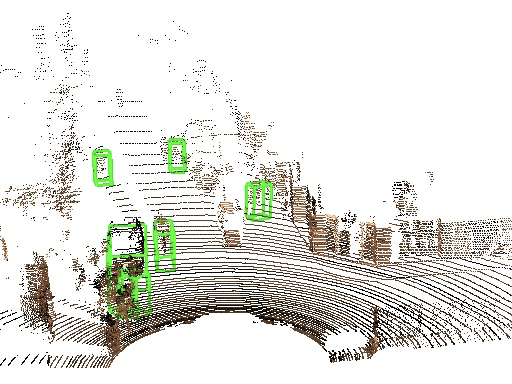}
        \includegraphics[width=\linewidth]{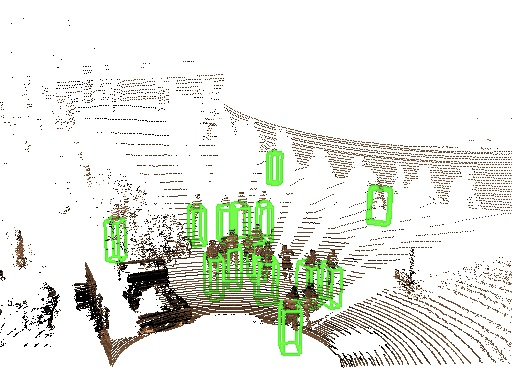}
        \includegraphics[width=\linewidth]{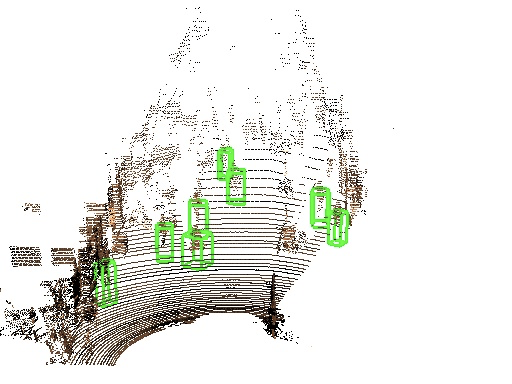}
        \caption{Ground Truth}
        \label{fig:semantickitti:appendix:gt:det}
    \end{subfigure}
    \caption{Qualitative comparisons between SECOND~\cite{yan2018second} and \cnnshort.}
    \vspace{-8pt}
    \label{fig:semantickitti:appendix:visualizations:det}
\end{figure*}

In \fig{fig:semantickitti:appendix:visualizations:det}, we demonstrate the superior performance of \cnnshort over SECOND~\cite{yan2018second}, which is a state-of-the-art single-stage 3D detector. Our \cnnshort has a large advantage in scenes with crowded small objects. In the first row of \fig{fig:semantickitti:appendix:visualizations:det}, our \cnnshort is capable of detecting a challenging small pedestrian instance missed by SECOND, and it also avoids duplicate predictions made by SECOND in the next three rows.

\end{document}